%% file: main.tex
\documentclass{bmvc2k}

\title{SELECT: SELEctive Context Transfer \textit{for} Class-Incremental Semantic Segmentation}

\addauthor{Avi Gupta}{avig@iiitd.ac.in}{1}
\addauthor{Saurabh Yadav}{yadavsaurabh@microsoft.com}{2,1}
\addauthor{Koteswar Rao Jerripothula}{kotesrj@iitk.ac.in}{3,1}
\addauthor{Tammam Tillo}{tammamtillo@tyust.edu.cn}{4,1}
\addinstitution{
IIIT Delhi\\
New Delhi, India
}
\addinstitution{
Microsoft, India
}
\addinstitution{
IIT Kanpur\\
Kanpur, India
}
\addinstitution{
TYUST, China
}

\runninghead{Gupta et al.}{Selective Context Transfer}
\input{preamble}
\newcommand*\colourcheck[1]{%
  \expandafter\newcommand\csname #1check\endcsname{\textcolor{#1}{\ding{52}}}%
}
\newcommand*\colourcross[1]{%
  \expandafter\newcommand\csname #1check\endcsname{\textcolor{#1}{\ding{55}}}%
}
\newcommand{\cross}{\textcolor{red}{\ding{55}}}

\usepackage{url}

\usepackage{graphicx}
\usepackage{booktabs}
\usepackage{hyperref}

\begin{document}

\maketitle

\input{sections/0_abstract}
\input{sections/1_intro}
\input{sections/2_related}
\input{sections/3_method}
\input{sections/4_experiments_new}
\input{sections/5_conclusion}
\input{sections/supplementary}

\bibliography{main}
\end{document}

%% file: preamble.tex
\usepackage{booktabs}       
\usepackage{amsfonts}       
\usepackage{nicefrac}       
\usepackage{microtype}      
\usepackage[table]{xcolor}
\definecolor{boxcolor}{HTML}{B85450}
\newcommand{\red}[1]{{\color{red}#1}}
\newcommand\crule[3][black]{\textcolor{#1}{\rule{#2}{#3}}}

\usepackage{graphicx}
\usepackage{booktabs}
\usepackage{multirow}
\usepackage{algorithm}
\usepackage{algpseudocode}
\usepackage{comment}
\usepackage{svg}
\usepackage{arydshln}
\usepackage{tikz}       
\usepackage{adjustbox}  
\usepackage{circledsteps}
\usepackage{amsmath}
\usepackage{wrapfig}
\usepackage{colortbl}
\usepackage{tabularray}
\usepackage{caption}
\usepackage{subcaption}
\usepackage{nicematrix}
\usepackage{amssymb}
\usepackage{bbm}
\usepackage{lipsum}
\usepackage{svg}
\usepackage{wrapfig}
\usepackage{pifont}
\usepackage{colortbl}

%% file: sections/0_abstract.tex
\begin{abstract}
Class-Incremental Semantic Segmentation (CISS) is fundamentally challenged by catastrophic forgetting and background shift, where learning new concepts degrades performance on previously seen classes. While existing methods attempt to balance stability (retaining old knowledge) and plasticity (learning new knowledge), they often fail to leverage prior knowledge effectively. These approaches typically rely on indiscriminate knowledge transfer or ambiguous initializations, which can dilute crucial semantic information. To overcome this limitation, we propose \textbf{\textit{SELECT\footnote{This paper is accepted in the proceedings of BMVC 2026}}}, a novel approach for \textbf{\underline{Sele}}ctive \textbf{\underline{C}}ontext \textbf{\underline{T}}ransfer, which instead grounds each new class in a small set of semantically similar past classes. Its core is a Context Transfer Attention mechanism that aggregates the learned tokens from similar classes into a structured initialization for the new class. To ensure this transfer does not corrupt the borrowed representations, we add a controlled noise perturbation and a margin-based context-transfer loss that enforces separation between the new class token and its source tokens. Extensive experiments on Pascal VOC and ADE20K show that SELECT consistently outperforms prior work, achieving mIoU of 2.2\% on VOC and 2.8\% on ADE, providing an effective handle on the stability-plasticity dilemma. Code
is available at \href{https://github.com/avigupta2798/SELECT}{https://github.com/avigupta2798/SELECT}.
\end{abstract}

%% file: sections/1_intro.tex
\section{Introduction}
\label{sec:intro}
Semantic segmentation, the task of assigning a class label to every pixel in an image, is a cornerstone of modern computer vision~\cite{shotton2006textonboost}.  It underpins critical vision applications, from navigating urban traffic to identifying pathological tissue \cite{DBLP:journals/tim/LiYSWLC23, DBLP:journals/imst/MourdiAS24}. However, most standard segmentation models are trained once and deployed as static systems. They lack the ability to actively learn and extend their representations when new object classes appear in a continuous stream.  Real-world systems, in contrast, require continual adaptation; integrating unfamiliar classes as they arise.  Instead of constantly retraining from scratch on continuously evolving data distributions, the focus should shift to Class-Incremental Semantic Segmentation (CISS), where the model sequentially learns new classes while retaining knowledge of previously seen ones.

The major bottleneck in CISS is catastrophic forgetting \cite{DBLP:journals/pami/LiH18a, kirkpatrick2017overcoming, DBLP:conf/cvpr/CaoZWYS0L024}, a challenging dynamic in which acquiring new knowledge disrupts the model's memory of previously learned classes. Mitigating this requires balancing three competing objectives. A model must maintain {\small\Circled{\textbf{1}}} \textit{\textbf{Stability}} to preserve past knowledge; {\small\Circled{\textbf{2}}} \textit{\textbf{Plasticity}} to absorb new class knowledge. Furthermore, it should facilitate {\small\Circled{\textbf{3}}} \textit{\textbf{Knowledge Transfer}}, intelligently using prior learned concepts to accelerate the learning of new ones. An effective CISS approach must strike a balance among these three challenges.

\begin{figure*}
    \centering
    \includegraphics[width=\textwidth]{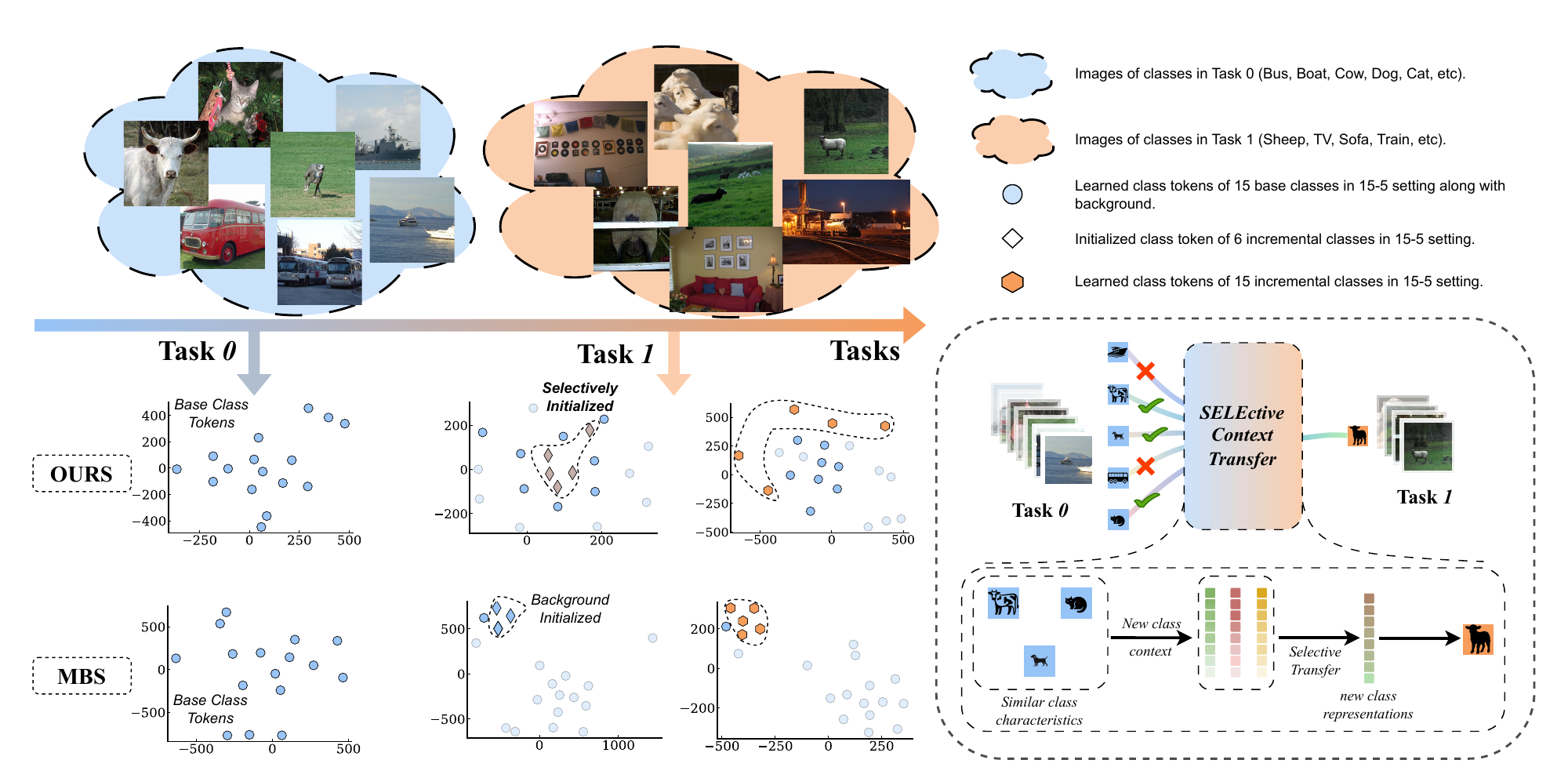}
    \vspace{0pt}
    \caption{\small \textit{Comparative analysis of knowledge transfer:} At Task 0, both our method and MBS~\cite{DBLP:journals/corr/abs-2407-11859} learn the initial classes identically. However, at Task 1, MBS initializes the new class tokens using the background. This causes the new tokens to cluster in an isolated corner, far from meaningful features. In contrast, our strategy dynamically anchors new classes to semantically similar past classes, resulting in a clean, well-distributed feature space. \textbf{[Bottom-Right]} \textit{Overview of \textbf{Sele}ctive \textbf{C}ontext \textbf{T}ransfer (\textbf{SELECT}):} Our approach selectively transfers useful knowledge from related past classes, allowing the model to efficiently learn new concepts without overwriting what it already knows.}
    \label{fig:teaser}
    \vspace{-10pt}
\end{figure*}

Most existing CISS methods strike this balance only partially. A common approach is to initialize new-class representations from the ``background'' class~\cite{DBLP:conf/cvpr/CermelliCD23, DBLP:conf/nips/ChaKYM21, DBLP:conf/wacv/GoswamiSWS23, DBLP:journals/corr/abs-2407-11859}. However, the background is a semantically ambiguous, noisy mixture of everything the model has not yet learned to recognize. Initializing a specific new class from this mixture creates a poor inductive bias, affecting the model's plasticity. As shown in Fig.~\ref{fig:teaser}, the background initialization in MBS~\cite{DBLP:journals/corr/abs-2407-11859} causes the initial representations to overlap and isolate from meaningful features, and they remain close to the background even after training. Another alternative works transfers knowledge from the past classes via global distillation~\cite{NEURIPS2020_d7488039, DBLP:conf/nips/Hiratani24}. While this captures broad context, it dilutes the few relevant classes with noise from many irrelevant ones.

To overcome these limitations, we present a CISS framework that moves away from background heuristics. To this end, we introduce \textbf{\textit{SELECT}} (\underline{SELE}ctive \underline{C}ontext \underline{T}ransfer). The core idea of SELECT is to first identify a small subset of past classes, $\mathcal{C}_s$, that are semantically closest to the incoming new class. This selective focus addresses dilution issues of global distillation. It then aggregates the learned tokens of these similar source classes into a single initialization through \textit{Context Transfer Attention (CTA)} mechanism. CTA selectively distills and modulates only the most relevant contextual features from the source classes in $\mathcal{C}_s$ to guide the learning of new ones. This provides a distinct, robust semantic initialization that directly promotes plasticity while ensuring efficient knowledge transfer.

Anchoring the new class to the old ones, however, risks disrupting the very representations it borrows from~\cite{incrementer}. If the new class token is placed too close to its sources, subsequent training can collapse its decision boundaries (validated in Table~\ref{tab:ablation_1}). To prevent this, we use a two-part approach. First, we add a small controlled noise perturbation to the transferred token. This small addition gives the newly initialized tokens some room to adapt to the new class while remaining semantically anchored. Second, a dedicated \textit{Context Transfer Loss} enforces a minimum margin between the new token and its source tokens, preventing representational overlap and retaining old knowledge. Together, CTA learns new classes while the margin loss preserves the old, balancing stability and plasticity.

Our key contributions are summarized as follows: {\small \Circled[fill color=black, inner color=white]{\textbf{1}}} we introduce a selection strategy that retrieves a small subset of semantically related past classes to support the learning of new classes; {\small \Circled[fill color=black, inner color=white]{\textbf{2}}} Context Transfer Attention (CTA), a mechanism that aggregates the most relevant prior tokens into a structured initialization for the new class; {\small \Circled[fill color=black, inner color=white]{\textbf{3}}} a Context Transfer Loss that enforces a margin between new and source classes, preventing the new class from overwriting old knowledge; and {\small \Circled[fill color=black, inner color=white]{\textbf{4}}} extensive experiments on Pascal VOC and ADE20K, where SELECT achieves state-of-the-art performance across most settings.

%% file: sections/2_related.tex
\section{Related Works}
\label{sec:related_works}

\textbf{Class-Incremental Semantic Segmentation: }Class-Incremental learning (CIL) has attracted growing attention for its ability to learn new knowledge without catastrophically forgetting the past~\cite{DBLP:journals/pr/WangSZXYL25, DBLP:conf/iclr/0003M24, DBLP:conf/eccv/KimLP24, kamra2017deep, shin2017continual, rosenfeld2018incremental}.  Some methods~\cite{DBLP:conf/iclr/HuangCH24, DBLP:journals/corr/abs-2303-13113} have used regularization-based approaches where the models introduce the regularizing terms and often penalize changes in the parameters to approximate the similar output distribution on previous tasks as the old model. Architectural-based methods~\cite{DBLP:conf/wacv/0005SDCWG24, DBLP:conf/aaai/BonatoPSN24, kirkpatrick2017overcoming} aim at updating the network modules that are adaptable to incremental tasks. Other approaches like replay-based \cite{DBLP:journals/access/LimZKL24, DBLP:journals/tip/ZhouYHZW24, lopez2017gradient, chaudhry2018efficient} memorize a small set of training images or representations from previous tasks and utilize them while learning the new tasks.

Advancements in CIL tasks encouraged us to move towards dense prediction tasks, such as semantic segmentation \cite{segmenter, segvit, xie2021segformer}, which remain susceptible to catastrophic forgetting. To alleviate such issues, \cite{DBLP:conf/iccvw/MichieliZ19} proposed class-incremental semantic segmentation that aims to segment the new classes without forgetting the previously learned ones. MiB \cite{DBLP:conf/cvpr/CermelliMB0C20} addresses the background shift problem using knowledge distillation. Following that, several works \cite{DBLP:conf/cvpr/MichieliZ21, DBLP:journals/corr/abs-2412-12669, DBLP:journals/corr/abs-2409-08516, DBLP:conf/wacv/GoswamiSWS23, DBLP:conf/cvpr/GongYWX24, DBLP:conf/cvpr/CermelliCD23, Kim_2024_CVPR, Zhang_2023_ICCV, DBLP:conf/eccv/ZhaoYFL22, chen2024strike} have been introduced to mitigate this issue. 
PLOP \cite{DBLP:conf/cvpr/DouillardCDC21} addressed this problem using multi-scale distillation and a pseudo-labelling strategy, while incrementer \cite{incrementer} proposed a transformer-based architecture. Our work resembles \cite{DBLP:conf/cvpr/CermelliMB0C20, DBLP:journals/corr/abs-2407-11859, DBLP:journals/corr/abs-2407-09838}, where we utilize the contrastive learning strategy to establish the difference between categories. However, our method differs in that we establish contrast between the tasks and the categories within each task.\\
\textbf{Knowledge Transfer in Incremental Learning: }Knowledge transfer strategies aim to propagate learned representations across diverse domains to enhance generalization~\cite{lopez2017gradient, chaudhry2018efficient, 10943898}. While early works~\cite{zeng2019continual, dhar2019learning, ruvolo2013ella, DBLP:conf/nips/Hiratani24} demonstrated the efficacy of knowledge reuse, they predominantly relied on traditional algorithms—such as linear regression~\cite{ruvolo2013ella}—where the phenomenon of catastrophic forgetting is inherently absent. Subsequent deep neural network approaches~\cite{NEURIPS2020_d7488039} attempted to navigate mixed sequences of similar and dissimilar tasks to mitigate forgetting, yet often struggled with representation interference. More recently, within the context of class-incremental semantic segmentation, \cite{DBLP:conf/eccv/XieLXWZL24} introduced a method to reuse priors from previously seen categories to initialize novel ones. However, as demonstrated in our empirical analysis (see $\S$\ref{sec:oursvsnest}), such uncalibrated transfer mechanisms prove representationally inefficient. To overcome this limitation, we propose a targeted semantic initialization strategy. By selectively identifying and retrieving only the most semantically aligned classes from the established knowledge base, we dynamically initialize the learned class embeddings within the decoder for the newly introduced categories.

%% file: sections/3_method.tex
\section{Methodology}
\label{sec:method}
\subsection{Problem Formulation}

We model the Class-Incremental Semantic Segmentation (CISS) problem as a sequence of distinct learning steps. Let $f$ represent an encoder-decoder architecture~\cite{segmenter} mapping an input image to dense pixel-wise predictions.

\textit{The learning process is a sequence of tasks indexed by $t \in \{0, 1, \dots, T\}$. At task $t$, the model receives dataset $\mathcal{D}_t = \{(x_i, y_i)\}_{i=1}^{N_t}$, where $x_i \in \mathbb{R}^{H \times W \times 3}$ and the ground-truth $y_i$ contains labels from a novel class set $\mathcal{C}_t$. The sets are mutually disjoint such that $\mathcal{C}_i \cap \mathcal{C}_j = \emptyset$ for any $i \neq j;\ i,j\in\{0, 1, \dots, T\}$. The objective at task $t > 0$ is to optimize the model $f_t$ for the cumulative label space $\mathcal{C}_{\leq t} = \bigcup_{i=0}^t \mathcal{C}_i$ without accessing historical data $\bigcup_{i=0}^{t-1} \mathcal{D}_i$. Our architecture relies on learnable class tokens.}

\begin{figure*}[t]
    \centering
    \includegraphics[width=\linewidth]{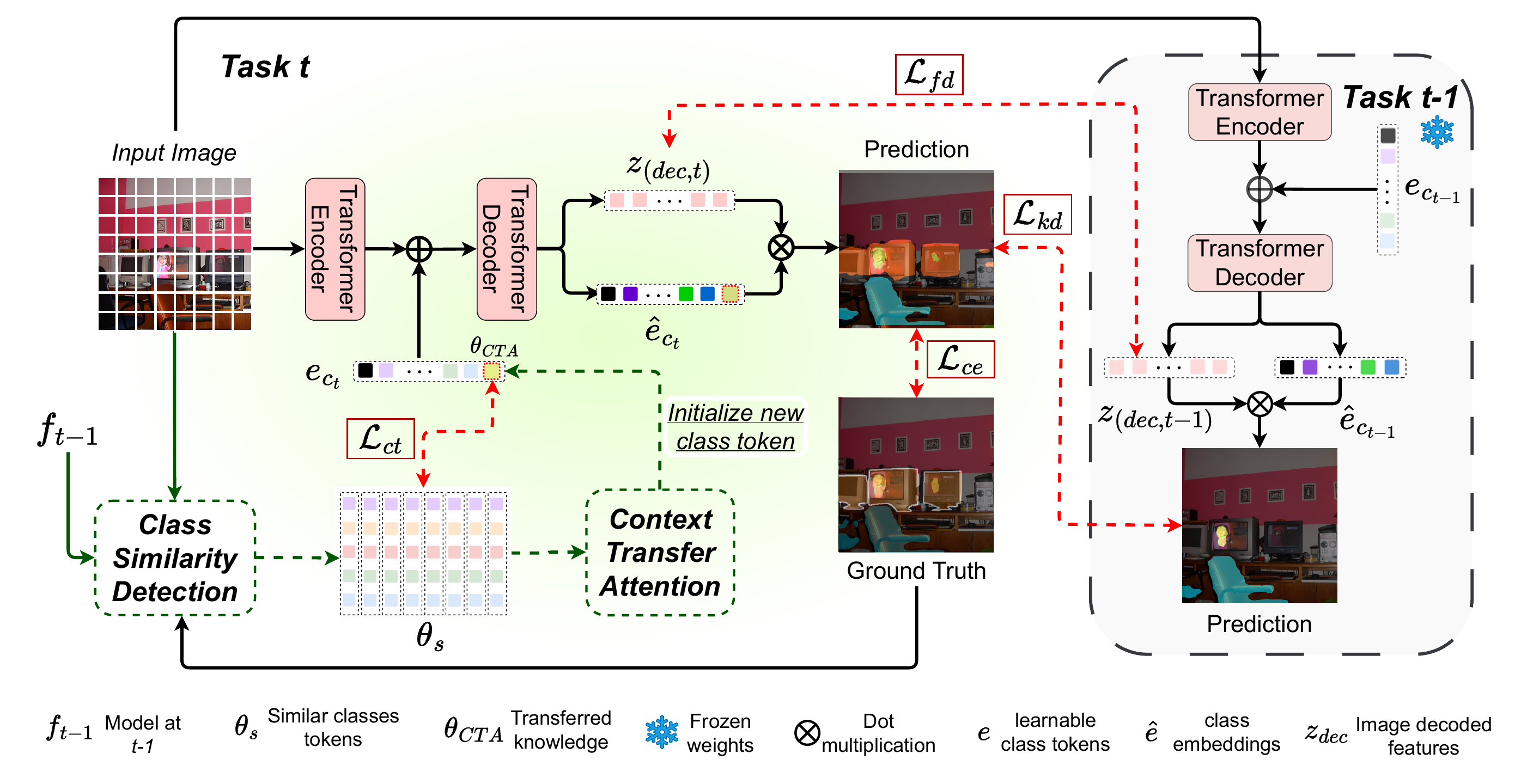}
    \vspace{0pt}
    \caption{\small \textbf{SELEctive Context Transfer (SELECT).} The task $t$ dataset is processed by the previous task's model $f_{t-1}$ to identify the most similar class for each image using \textbf{Class Similarity Detection} (see $\S$\ref{sec:csd}). 
    The similar classes' tokens $\theta_s$ are then processed by \textbf{Context Transfer Attention} (see $\S$\ref {sec: cbrs}) to formulate the adaptive token $\theta_{CTA}$ for the new class that is further used in the current task's model $f_t$. The model is trained using $\mathcal{L}_{ce}, \mathcal{L}_{fd}, \mathcal{L}_{kd}$, and the proposed context transfer loss $\mathcal{L}_{ct}$ enforcing distinction between $\theta_{CTA}$ and $\theta_{s}$ (see $\S$\ref{sec:obj})}.
    \label{fig:architecture}
    \vspace{-10pt}
\end{figure*}

At $t=0$, $f_0$ is trained on an initial set of classes $\mathcal{C}_0$. Specifically, for an input $x$, we obtain the encoded feature representations as $z^{enc} = f_0^{enc}(x)$.
Alongside these encoded representations, we introduce a set of learnable class tokens $e_{\mathcal{C}_0}\in\mathbb{R}^{\mathcal{C}_0\times d}$, where $d$ is the token dimension. These tokens, one for each class, are randomly initialized and optimized to capture the core semantic characteristics of their respective classes, which are then passed into the decoder (alongside encoded features) for segmentation: $[z^{dec},\hat{e}_{\mathcal{C}_0}] = f_0^{dec}(z^{enc},e_{\mathcal{C}_0})$. The predicted mask $\hat{y}$ is obtained via dot product of decoded spatial features $z^{dec}$ and the image-specific class representations: $\hat{y} = z^{dec} \otimes \hat{e}_{\mathcal{C}_0}$. At $t>0$, we initialize the new task model $f_t$ with the previous model $f_{t-1}$ and freeze $f_{t-1}$. Unlike the base task ($t=0$), the learnable class token at task $t>0$, $e_{c_{new}}$, for a new class $c_{new}\in \mathcal{C}_t$, is not randomly initialized, but selectively curated using previously learned class tokens. The goal is to produce a model $f_t$ that performs well not only on the new classes $\mathcal{C}_t$ but also on the cumulative set of all classes seen so far, $\mathcal{C}_{\leq t}=\bigcup_{i=0}^t \mathcal{C}_i$. 
Specifically, when $c_{new}$ is introduced, the framework first identifies previously learned classes in $\mathcal{C}_{0:t-1}$, where $\mathcal{C}_{0:t-1}$ is the set of all classes up to task $t-1$, that might be semantically similar. Knowledge is then selectively transferred from related classes to construct an informed, guided initial representation for the new class. This transferred knowledge provides a strong inductive bias, grounding the new concept within the model's existing semantic space. Finally, this new representation is fine-tuned using the current task's data $\mathcal{D}_t$. 
An overview of the proposed \textbf{\textit{SELECT}} framework is visualized in Fig.~\ref{fig:architecture}.

\begin{figure*}[h]
\centering
\includegraphics[width=\textwidth]{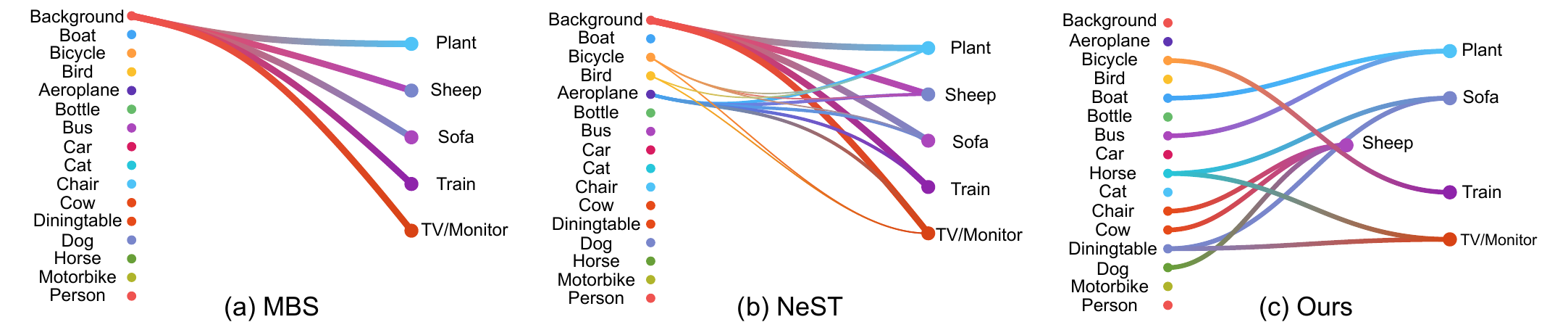}
\vspace{0pt}
\caption{\small Visual representation of knowledge transfer in different approaches. (a) In \textit{MBS}~\cite{DBLP:journals/corr/abs-2407-11859}, the initial knowledge for incremental classes is transferred entirely from the background; (b) In \textit{NeST}~\cite{DBLP:conf/eccv/XieLXWZL24}, most of the initial knowledge for incremental classes is transferred from the background; (c) In \textbf{\textit{ours}}, the focus is on semantically relevant classes.}
\label{fig:analysis2}
\vspace{-5pt}
\end{figure*}

\subsection{Analysis on Efficient Knowledge Transfer}
\label{sec:oursvsnest}

In this section, we analyze the predominant knowledge transfer strategies used in prior CISS work and highlight their underlying limitations. Our analysis is grounded in two standard incremental settings on the Pascal VOC dataset, $15-1$ and $15-5$, under the overlapped scenario detailed in $\S$\ref{sec:setup}.

\textit{Initializing a new class token without a clear semantic anchor, such as using random vectors, or the undefined background, leaves the model without targeted guidance. This ambiguity maximizes initial gradient variance, further hurting convergence.}

A model's convergence heavily depends on where it starts. Current methods generally rely on three common initialization strategies:
1) \textit{Random initialization:} Starting with random vectors offers no meaningful anchor~\cite{DBLP:conf/cvpr/PhanTPTB22}.
2) \textit{Background initialization:} The background subspace ($\mathcal{F} \setminus \mathcal{C}_{\leq t-1}$) is a catch-all for undefined objects.~\cite{DBLP:journals/corr/abs-2407-11859, DBLP:conf/cvpr/CermelliMB0C20}.
3) \textit{Flawed Distillation:} Attempting to reuse past classes often accidentally selects features that have collapsed back into the background noise~\cite{DBLP:conf/eccv/XieLXWZL24}. 

\input{Tables/nest}Ultimately, all three approaches provide a vague starting point ($e_{c_{new}}$) rather than a sharp, low-variance direction. As shown in Fig.~\ref{fig:loss} (a-b), the ambiguity in initialization in BARM, MBS, and REMINDER hinders convergence and performs poorly overall. Table~\ref{tab:nest} highlights this issue specifically for NeST~\cite{DBLP:conf/eccv/XieLXWZL24}. 
NeST pre-tunes new classes by nominally transferring from prior classes. NeST$_{bg}$ is our controlled variant with background-only initialization; both were run in the same environment. Table~\ref{tab:nest} on short 15-5 setting, shows class-based initialization offers only a marginal (+2.1 All) gain over bg, while on long 15-1, NeST falls 6.9 below its own bg variant. Hence, NeST’s selected representations are progressively dominated by background noise (motivating selective transfer) (as visualized in Fig.~\ref{fig:analysis2}). This proves that simple regularization isn't enough. Instead, our method completely bypasses these ambiguous spaces by leveraging prior classes' learned representations, providing a highly structured and reliable initialization.

\subsection{\underline{\textit{SELE}}ctive \underline{\textit{C}}ontext \underline{\textit{T}}ransfer (SELECT)}
To avoid using the ambiguous initialization strategies, we introduce \textit{SELECT} (Fig.~\ref{fig:architecture}). When a new class ($c_{new} \in \mathcal{C}_t$) arrives, we do not start from scratch or from the background. Instead, we initialize it using a carefully chosen subset of related past classes, denoted as $\mathcal{C}_s$. In $\S$~\ref{sec:csd}, we will formulate the strategy to selectively identify the set of those similar classes. Finally, in $\S$~\ref{sec: cbrs}, we will initialize the new class using the identified classes.
\input{sections/3_b_class_similarity}
\input{sections/3_a_cta}
\input{sections/3_c_obj}

%% file: Tables/nest.tex
\begin{wrapfigure}{l}{0.3\linewidth}
\vspace{-10pt}
\centering
\captionof{table}{\small Performance comparison of NeST \cite{DBLP:conf/eccv/XieLXWZL24} on Pascal VOC. $\dagger$ denotes results are reproduced. $bg$ denotes that only background initialization is used. \crule[red]{.18cm}{.18cm} denotes the difference in the performance.}
\label{tab:nest}
\vspace{5pt}
\setlength{\tabcolsep}{2pt}
\renewcommand{\arraystretch}{1}
\resizebox{.32\textwidth}{!}{
\centering
\begin{NiceTabular}{l|ccc|ccc}
\toprule
\multirow{2}{*}{}  & \multicolumn{3}{c|}{15-1}   & \multicolumn{3}{c}{15-5}                        \\  
                      & 0-15     & 16-20      & All     &  0-15     & 16-20     & All      \\ \midrule
    \Block[tikz={shade, shading=radial, inner color=lime!80, outer color=white}]{1-1}{}{NeST$^\dagger$}    & 57.2      & 32.1      & 51.3    & \Block[tikz={shade, shading=radial, inner color=lime!80, outer color=white}]{1-3}{} 76.8      & 52.6      & 71.0         \\
    
   \Block[tikz={shade, shading=radial, inner color=orange!80, outer color=white}]{1-1}{}{NeST$^\dagger_{bg}$}   & \Block[tikz={shade, shading=radial, inner color=orange!80, outer color=white}]{1-3}{}65.6 & 34.6 & 58.2 
    & 75.2      & 49.0      & 68.9        \\ \midrule
                      & \textbf{\red{-8.4}}      & \textbf{\red{-2.5}}      & \textbf{\red{-6.9}}     & \textbf{\red{+1.3}}      & \textbf{\red{+3.6}}      & \textbf{\red{+2.1}}        \\ \bottomrule
\end{NiceTabular}
}
\end{wrapfigure}

%% file: sections/3_b_class_similarity.tex
\subsubsection{Identifying Similar Classes}
\label{sec:csd}
In this section, we identify the set of similar old classes, $\mathcal{C}_s \subseteq \mathcal{C}_{0:t-1}$, in a new task $t$, for knowledge transfer.

\textit{Old class ($c_{old} \in \mathcal{C}_{0:t-1}$) is semantically relevant to a new class ($c_{new} \in \mathcal{C}_t$) if the established representation for $c_{old}$ ($e_{c_{old}}$) exhibits stability when they are exposed to the masked images of the new class.}

\begin{wrapfigure}{l}{0.48\linewidth}
\centering
\vspace{-15pt}
\includegraphics[width=0.42\textwidth]{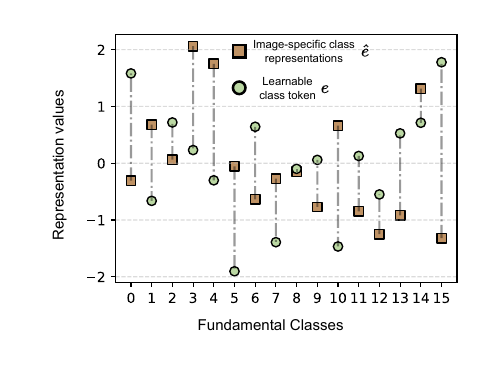}
\vspace{-5pt}
\caption{\small Representational perturbation between $e_{\mathcal{C}_{0:t-1}}$ and $\hat{e}_{\mathcal{C}_{(0:t-1)}}$.}
\label{fig:analysis7}
\vspace{0pt}
\end{wrapfigure} Specifically, model $f_{t-1}$ encapsulates the learned knowledge of all past classes $\mathcal{C}_{0:t-1}$. We use this frozen model as a probe. For each image-mask pair $(x, y_{c_{new}}\in \mathcal{D}_t)$, for a new class, we feed the masked image ($\mathcal{I}_{c_{new}} = x\odot y_{c_{new}}$) into $f_{t-1}$, where $\odot$ is the Hadamard product, to get the fixed learned tokens $e_{\mathcal{C}_{0:t-1}}$ and what the decoder predicts, when it infers on the masked image, $\hat{e}_{\mathcal{C}_{(0:t-1)}, \mathcal{I}_{c_{new}}}$. The GT masks $y_{c_{new}}$ are the current task’s training annotations and are used by every supervised CISS approach; SELECT requires no additional supervision. Finally, we measure the semantic relevance, over the entire dataset $\mathcal{D}_t$, by assessing the distance between $e_{\mathcal{C}_{0:t-1}}$ and $\hat{e}_{\mathcal{C}_{(0:t-1)}, \mathcal{D}_t}$. We calculate the semantic distance by taking the Euclidean norm between them:
\begin{equation}
\text{dist}(\hat{e}, e\ |\  \mathcal{D}_t) = \| \hat{e}_{\mathcal{C}_{(0:t-1)}, \mathcal{D}_t} - e_{\mathcal{C}_{0:t-1}} \|_2
\end{equation}
Through $\text{dist}(\hat{e}, e\ |\  \mathcal{D}_t)$, we identify the most similar class for each image in $\mathcal{D}_t$.

\textit{To prove this hypothesis, we perform an empirical analysis. We train the base-class model $f_0$ on 15 unique classes. In the incremental setting, we introduce a new class, ``sheep'', and feed masked sheep images into the base model. Fig.~\ref{fig:analysis7} shows the average representation over all sheep images; we observe that some base classes exhibit stronger semantic relevance than others, with significantly lower distances.}

We form the final set of similar classes, $\mathcal{C}_s$, by counting how often each class is selected across the dataset, and retaining only those whose frequency exceeds a predefined threshold. The frequency threshold is controlled by a hyperparameter $\varepsilon \in [0,\ 1]$. The resulting set $\mathcal{C}_s$ contains past classes that provide robust ``positive contextual knowledge''. Similar plots targeting individual masked sheep images are provided in the supplementary file.

%% file: sections/3_a_cta.tex
\subsubsection{Context-driven Selective Knowledge Transfer}
\label{sec: cbrs}
Having identified the set of semantically similar past classes $\mathcal{C}_s$ (from \S\ref{sec:csd}), and aggregating their learned tokens $\theta_s = \{e_c\}_{c\in\mathcal{C}_s}$, we formulate \textbf{C}ontext \textbf{T}ransfer \textbf{A}ttention (CTA) for transferring their knowledge. 

\textit{By building a new learnable token as a convex combination of past ones, we force the model to start in a meaningful, already-learned area of the feature space. This drastically shrinks the search space compared to starting from scratch with an ambiguous initialization.}

We compute a new guided token $\theta_{CTA} \in \mathbb{R}^D$ that captures the characteristics of $\theta_s$ using an attention mechanism. The collective context of similar classes serves as the query, encapsulating the gist of the semantic neighborhood. This combined information is then used to attend to individual learned tokens, as key, in $\theta_s$ and generate attention scores. The attention scores are finally multiplied by the individual learned tokens' values to obtain the weighted information from all learned similar classes' tokens.

\begin{equation} 
\theta_{CTA} = \text{softmax}\Bigg( \frac{\Big(\frac{1}{n}\sum_{j=1}^{n}\theta_{s}^{j}\Big) \times (\theta_{s})^{\textbf{T}}}{\sqrt{d}} \Bigg) \theta_{s} 
\label{eq:cta_attention}
\end{equation}
Here $d$ is the embedding dimension, $(\cdot)^{\textbf{T}}$ denotes the transpose, and $n$ is the total number of similar classes in $\theta_s$. The softmax term produces a vector of attention weights that assign importance to each source class, enabling selective knowledge transfer.

\paragraph{Enforcing Representational Separation via Perturbation.}
While Eqn.~\ref{eq:cta_attention} provides a strong semantic prior, it poses a challenge: 

\textit{If the new starting token $\theta_{CTA}$ is placed too close to the old tokens in $\theta_s$, training on the new data will cause their representations to merge. This overlap destroys the decision boundaries of the old classes in $\mathcal{C}_s$, leading to catastrophic forgetting or misclassifying old classes as new ones during inference.}

We empirically validate this in Table~\ref{tab:ablation_1}, which shows that initializing the new learnable tokens solely using $\theta_s$ results in a sharp drop in accuracy of corresponding old classes. To mitigate this, we introduce a controlled perturbation into $\theta_{CTA}$. Specifically, instead of directly using $\theta_{CTA}$, we regularize it by interpolating with a noisy version of itself, thereby creating a buffer zone in the feature space. This gives the new class the ability to adapt without confusing it with the old classes, while maintaining semantic alignment. We calculate the final token as:
\begin{equation}
\label{eq:noise}
    \hat{\theta}_{CTA} = \alpha\theta_{CTA} + (1-\alpha)\mathcal{N}(\theta_{CTA}, \sigma^{2})
\end{equation}
Here, $\mathcal{N}(\theta_{CTA}, \sigma^{2})$ is a sample from a Gaussian distribution with mean $\theta_{CTA}$ and variance $\sigma^2$. The hyperparameter $\alpha\in [0,1]$ controls the trade-off between preserving the precise transferred knowledge (alignment) and introducing diversity (separation). 

Hence, for a new class $c_{new}\in\mathcal{C}_t$, we initialize the new class token $e_{c_{new}}$ with $\hat{\theta}_{CTA}$. 

%% file: sections/3_c_obj.tex
\subsection{Objective Function}
\label{sec:obj}

Our base training setup follows standard CISS methods. We use a standard cross-entropy loss ($\mathcal{L}_{ce}$) for learning new classes, along with feature ($\mathcal{L}_{fd}$) and knowledge ($\mathcal{L}_{kd}$) distillation losses to prevent the model from forgetting what it already knows.\\
\textbf{Cross-Entropy Loss ($\mathcal{L}_{ce}$): }The cross-entropy loss is used in both base and incremental tasks. In the base task ($t=0$), we use the conventional cross-entropy loss between the predicted mask $\hat{y}_{0}\in \mathbb{R}^{H\times W\times (0:\mathcal{C}_t)}$ and its corresponding ground truth $y_{t}\in \mathbb{R}^{H\times W}$:
\begin{equation}
    \mathcal{L}_{ce}(y_{0}, \hat{y}_0) = -\frac{1}{HW}\sum_{i=1}^{HW} y_{0, i}\log \hat{y}_{0, i}
\end{equation}
where $H$ and $W$ are the height and width of an image. In the incremental task ($t>0$), we use the previous task's predicted label along with the ground truth of the current class as the pseudo label $\tilde{y}_{t}(y_{t}, \hat{y}_{t-1})$ and calculate the loss with the combined predicted label $\hat{y}_{t}$. The updated $\mathcal{L}_{ce}$ is:
\begin{equation}
    \mathcal{L}_{ce}(\tilde{y}_{t}, \hat{y}_{t}) = -\frac{1}{HW}\sum_{i=1}^{HW} \tilde{y}_{t, i}(y_{t}, \hat{y}_{t-1})\log \hat{y}_{t, i}
\end{equation}
\textbf{Feature Distillation Loss ($\mathcal{L}_{fd}$): }We employ this loss to prevent the current model's features from deviating from the previous ones. This loss is calculated between the output feature patches from the decoders of both the previous $z^{dec}_{t-1}$ and the current $z^{dec}_t$ models.
\begin{equation}
    \mathcal{L}_{fd} = \frac{1}{HW}\sum_{i=1}^{HW}\|z^{dec}_{t-1,i} - z^{dec}_{t,i}\|^{2}
\end{equation}
\textbf{Knowledge Distillation Loss ($\mathcal{L}_{kd}$): }We employ this loss to distil the previous model's prediction. This loss is calculated between the output predictions of both the previous $\hat{y}_{t-1}$ and current models $\hat{y}_{t}$.
\begin{equation}
    \mathcal{L}_{kd} = -\frac{1}{HW}\sum_{i=1}^{HW}\hat{y}_{t-1,i}\log \hat{y}_{t,i}
\end{equation}

However, our selective initialization brings the new class token ($\hat{\theta}_{CTA}$) closer to the source tokens ($\theta_{s}$), thereby increasing the risk of misclassification. To counteract this, we introduce a \textbf{\textit{Context Transfer Loss}} ($\mathcal{L}_{ct}$), as a margin-based penalty:
\begin{equation}
\label{eq:lct}
    \mathcal{L}_{ct} = \frac{1}{|\mathcal{C}_s|}\sum_{c \in \mathcal{C}_s}\max(0, M - \|e_{c_{new}} - e_c\|_2)
\end{equation}

\textit{By training the network to push this loss to zero, we encourage that the distance between the new class token and its source tokens stays larger than the margin $M$. This keeps the representations clearly separated.
}

If the loss reaches its minimum ($\mathcal{L}_{ct} = 0$), the maximum function dictates that every individual term inside the sum must also be zero. For that to happen, $M - \|e_{c_{new}} - e_c\|_2 \leq 0$ for every source class $c \in \mathcal{C}_s$. Rearranging this simply gives $\|e_{c_{new}} - e_c\|_2 \geq M$. In plain terms, this loss forces the new token to remain at least $M$ away from any source token, actively protecting the boundaries of the old classes. The final training objective is formulated as $\mathcal{L}_{total} = \mathcal{L}_{ce} + \lambda_{kd}\mathcal{L}_{kd} + \lambda_{fd}\mathcal{L}_{fd} + \lambda_{ct}\mathcal{L}_{ct}$.

%% file: sections/4_experiments_new.tex
\section{Experiments}
\label{sec:experiments}
\subsection{Experimental Setup}
\label{sec:setup}
\textbf{Training setting: }We evaluate our proposed approach under two distinct experimental scenarios -- \textit{disjoint} and \textit{overlapped} \cite{DBLP:journals/corr/abs-2407-11859, DBLP:journals/corr/abs-2407-09838}. In both scenarios, ground-truth labels are provided exclusively for the classes designated for the current task $t$. The primary distinction lies in the composition of the input images. In the standard disjoint setting, images in the current task's dataset $\mathcal{D}_{t}$ contain only instances from classes seen so far, $\mathcal{C}_{1:t}$. Conversely, the overlapped setting permits images to also contain unlabeled instances of objects from future classes $\mathcal{C}_{1:T}$. Hence, the overlapped scenario more closely aligns with real-world data streams, presenting a significantly more challenging and realistic benchmark.\\
\textbf{Datasets and Evaluation Metrics: }Following prior works \cite{DBLP:journals/tip/WangWYZH24, DBLP:journals/corr/abs-2405-09858, DBLP:journals/corr/abs-2407-09838, DBLP:journals/corr/abs-2407-11859}, we conduct our experiments on two public benchmarks: Pascal VOC 2012 \cite{DBLP:journals/ijcv/EveringhamEGWWZ15} and ADE20K \cite{DBLP:conf/cvpr/ZhouZPFB017}. The Pascal VOC dataset comprises 10,582 training and 1,449 testing images across 20 foreground categories. ADE20K is a larger-scale dataset with 150 classes, containing 20,210 training and 2,000 testing images. 
For the Pascal VOC, we evaluate on the 15-1, 15-5, and 19-1 splits under both disjoint and overlapped scenarios. The 15-1 split, for example, consists of a base task with 15 classes, followed by 5 incremental tasks, each adding a single new class, for a total of 6 tasks. For ADE20K, we use the 100-50, 50-50, 100-10, and 100-5 splits, focusing on the more realistic overlapped scenario. We measure the performance using mean intersection-over-union (mIoU) for the base classes (old), incremental classes (new), and all classes combined (all). “all” is computed as the mean over classes with non-zero IoU, including background, while class-groups (1-15, 16-20) are means over their fixed class sets, including zero-IoU classes and excluding background. \\
\textbf{Implementation Details: }Our approach is built upon a ViT-B/16 backbone \cite{dosovitskiy2020image}, pre-trained on ImageNet. The segmentation head is a transformer-based decoder inspired by the Segmenter \cite{segmenter}. Our training protocol largely follows the configuration in \cite{DBLP:journals/corr/abs-2407-11859}. We use the SGD optimizer and employ random rotation and cropping for augmentations. All experiments are implemented in PyTorch and conducted on a single workstation with a NVIDIA A100 GPU. For Pascal VOC, we set the initial learning rate to 1e-3 and train for 32 epochs with a batch size of 16. For ADE20K, we use a learning rate of 5e-4 and train for 64 epochs with a batch size of 8. We set the model hyperparameter $\alpha=0.9$ and $\sigma=0.05$. The threshold value $\varepsilon$ is fixed at 0.15. For our proposed loss $\mathcal{L}_{ct}$, we set the margin M at 1.0, and $\lambda_{ct}$ to 0.8. $\lambda_{kd}$, and $\lambda_{fd}$ are adopted from \cite{DBLP:journals/corr/abs-2407-11859}. Additional details are provided in the supplementary file.
\input{Tables/pascal_ade_final}

\subsection{Experimental Results \& Further Analysis}
\label{sec:pascal_results}
\textbf{Comparison with State-of-the-Art:} Table~\ref{tab:pascal_ade} details how our approach performs on the Pascal VOC and ADE20K datasets across various training setups. As shown, our method significantly outperforms existing baselines \cite{DBLP:journals/corr/abs-2407-09838, incrementer, DBLP:conf/cvpr/CermelliMB0C20, DBLP:conf/nips/ChaKYM21, DBLP:conf/eccv/XieLXWZL24, DBLP:journals/corr/abs-2407-11859} by a wide margin in nearly every setting. This is especially evident when comparing our model to \cite{DBLP:conf/eccv/XieLXWZL24}, a recent method that also leverages prior knowledge. We consistently outperform them across the board. Because \cite{DBLP:conf/eccv/XieLXWZL24} originally uses a Swin-B backbone, we also provide a direct, fair comparison using the ViT and CNN-based backbones, along with additional results on VOC and ADE20K, in the supplementary file.

We see some strong results in the 15-5 setting. This is a difficult setup, as the model learns multiple new classes simultaneously with highly varied data distributions. Despite this added complexity, our approach surpasses prior work by a significant margin. ADE20K features highly diverse and realistic scenes, and the overall scores reflect that difficulty compared to Pascal VOC. However, our proposed approach maintains a balance between the performance of the old and new classes. In our results, the performance gap between the base classes and the newly added incremental classes is significantly smaller than the margins reported in NeST \cite{DBLP:conf/eccv/XieLXWZL24}. This proves that our strategy for identifying and using similar past classes is much more effective at maintaining the delicate trade-off between stability and plasticity. 

\begin{figure*}[ht]
\centering
\includegraphics[width=\textwidth]{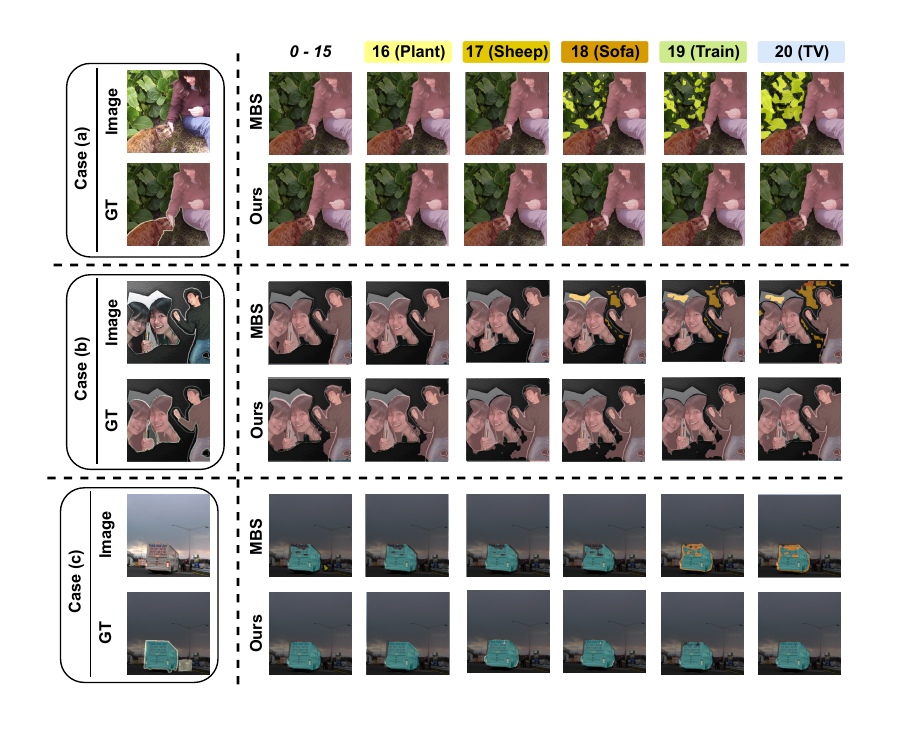}
\caption{\small Visual comparison on 15-1 scenario of the Pascal VOC between MBS and ours.}
\label{fig:pascal_vis}   
\end{figure*}

\begin{figure*}[ht]
    \centering
    \includegraphics[width=.9\textwidth]{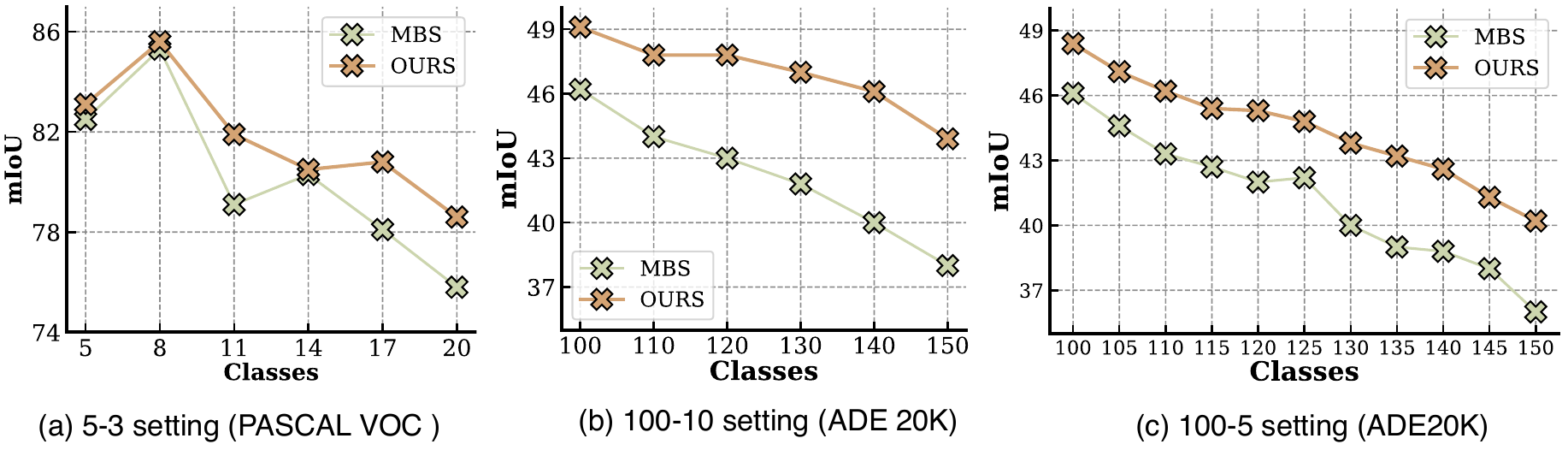}
    \vspace{10pt}
    \caption{\small Comparative analysis over long task sequences. (a) PASCAL VOC 5-3 (6 steps) setting; (b) ADE20K 100-10 (6 steps) setting; (c) ADE20K 100-5 (11 steps) setting.}
    \label{fig:longerseq}
\end{figure*}

We compare the stability of SELECT across longer task sequences with MBS. We adopt the ADE20K dataset with the 100-5 (11 steps) and 100-10 (6 steps) settings and the PASCAL VOC dataset with the 5-3 (6 steps) setting. As seen in Fig.~\ref{fig:longerseq}, SELECT consistently performs better than MBS across the incremental sequence. Specifically, in the 100-5 setting, we observe that SELECT's rate of performance decline is lower than MBS's, leading to greater stability across tasks. The visual results on Pascal VOC dataset are shown in Fig.~\ref{fig:pascal_vis}.

\begin{figure*}[ht]
\centering
    \includegraphics[width=\textwidth]{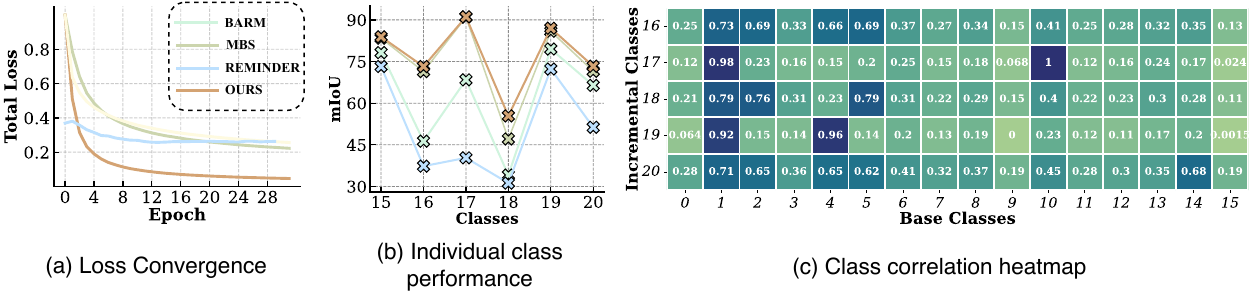}
    \vspace{5pt}
    \caption{\small (a-b) Comparative analysis of loss convergence and task performance on BARM, MBS, REMINDER, and Ours (SELECT) on 15-5 setting of PASCAL VOC dataset; (c) Correlation heatmap on PASCAL-VOC 15-5 setting between current and previous task classes.}
    \label{fig:loss} 
\end{figure*}

\noindent \textbf{Training Time and Convergence:} 
To evaluate effective convergence, we compare the loss of MBS, REMINDER~\cite{DBLP:conf/cvpr/PhanTPTB22}, and BARM~\cite{DBLP:journals/corr/abs-2407-09838} with our proposed \textit{SELECT}, in step 1 of the VOC 15-5 incremental setting (shown in Fig.~\ref{fig:loss}(a)). All methods show a decline in the early epochs, indicating less stable optimization. Conversely, SELECT consistently achieves better convergence, demonstrating superior training stability and robustness. To further assess, we also visualize, in Fig.~\ref{fig:loss}(b), the average performance of fundamental classes and the individual performance of 5 incremental classes. SELECT consistently performs better than the others.

\noindent \textbf{Effectiveness of class similarity $\mathcal{C}_s$:} In this study, we evaluate the effectiveness of using previously learned similar classes. In Fig.~\ref{fig:loss}(c), we visualize the correlation heatmap, on Pascal-VOC 15-5 setting, to assess the similarity of the current task's classes with the previously learned ones. We observe that rather than relying on a single, arbitrary class, incremental \input{Tables/ablation_sim_dissim}classes are selectively similar to a few. To further validate our hypothesis, we transfer knowledge from previous dissimilar classes ($\mathcal{C}_{ds} = \mathcal{C}_{0:t-1} - \mathcal{C}_s$) instead of the similar classes. We perform this ablation on the Pascal dataset with 15-1 and 15-5 settings. As seen in Table~\ref{tab:ablation_simivsdissimi}, $\mathcal{C}_{ds}$ significantly impacts the performance of both fundamental and incremental classes by a significant margin. Since dissimilar classes contain entirely different contextual information, their representations fail to provide meaningful information. These results confirm that similar class representations provide a more effective foundation, enabling the model to adapt to new classes while preserving knowledge of previously learned ones.\\

\noindent \textbf{Effectiveness of similarity metrics:} We perform this study to analyze the effectiveness of \input{Tables/ablation_similarity_metrics}using Euclidean distance as an efficient metric to identify the similarity between learned class tokens and image-specific class representations. A small distance implies a high semantic similarity. While other metrics like cosine similarity focus on orientation, Euclidean distance considers both magnitude and orientation, providing an appropriate measure of geometric distance in the latent space. As observed in Table~\ref{tab:ablation_3}, the Euclidean distance consistently outperforms the other metrics, particularly for incremental tasks.\\

\input{Tables/ablation_init_strat} \noindent \textbf{Analysis on Initialization Strategy: }As observed in Table~\ref{tab:ablation_init}, we present the analysis on two orthogonal contributions of SELECT, the CTA initialization ($\theta_{CTA}$) and the Context Transfer Loss ($\mathcal{L}_{ct}$), against a background-initialized baseline ($\theta_{bg}$) under the same settings. Replacing background initialization with $\theta_{CTA}$ while keeping the loss identical yields improvements of $\approx$9\% on 15-1 and $\approx$1\% on 15-5. Using a new class token from semantically similar past class representations, $\theta_{CTA}$ provides the decoder with a structured, low-variance starting point before any gradient update. The background token, by contrast, is a noisy representation of everything the model has not yet learned to recognize. Using the full loss with $\mathcal{L}_{ct}$ yields an additional $\approx$4\% on 15-1 and $\approx$2\% on 15-5. This confirms that, even with selective initialization, $\theta_{s}$ still creates proximity risk, leading to representation interference. $\mathcal{L}_{ct}$ resolves this by enforcing a minimum margin $M$ between $e_{c_{new}}$ and the similar source tokens.

\input{Tables/ablation_loss}\noindent \textbf{Effectiveness of loss components:} To quantify the contribution of each loss component ($\mathcal{L}_{ce},\ \mathcal{L}_{fd},\ \mathcal{L}_{kd},\ \mathcal{L}_{ct}$) within our CISS framework, we conduct a detailed study across 15-1 and 15-5 scenarios. As shown in Table~\ref{tab:ablation_2}, removing the feature distillation loss ($\mathcal{L}_{fd}$) or the knowledge distillation loss ($\mathcal{L}_{kd}$) leads to a catastrophic decline in performance, particularly on incremental classes. In the 15-1 setting, ablating $\mathcal{L}_{fd}$ causes the mIoU on new classes to plummet from 72.0\% to 14.2\%. Similarly, removing $\mathcal{L}_{kd}$ drops the same metric to 32.6\%. This underscores that $\mathcal{L}_{fd}$ is vital for learning high-quality, patch-level representations of novel classes, while $\mathcal{L}_{kd}$ is indispensable for mitigating catastrophic forgetting of the base classes. Our proposed loss provides a substantial performance boost. Removing $\mathcal{L}_{ct}$ degrades the overall mIoU from 80.5\% to 76.1\% in the 15-1 setting, with the most significant damage inflicted on the incremental classes (72.0\%$\rightarrow$60.7\%).
\\

\input{Tables/ablations_four}\noindent \textbf{Effectiveness of knowledge transfer:} In Table~\ref{tab:ablation_1}, we evaluate the class-wise performance to analyze the most efficient approach for knowledge transfer. Specifically, we investigate how transferring knowledge from previously learned similar classes affects the performance of all the classes. To this end, we compare our proposed attention-based approach~$\theta_{CTA}$ to two major baselines \textit{(i)} $\theta_{Best}$ (selects the most similar class from the subset) and \textit{(ii)} $\theta_{Avg}$ (averages the representations of all similar classes). Additionally, we also evaluate the impact of integrating controlled noise~$\mathcal{N}$ across the strategies.
Primarily, we observe that our proposed approach consistently outperforms the baselines by a significant margin. In particular, our attention-based framework achieves at least 11\% and 41\% improvement over the baselines on the 1-15 and 16-20 tasks, respectively. Upon looking closely, in $\theta_{Best}$ and $\theta_{Avg}$ methods, transferring knowledge from certain classes (\textit{e.g.} aero) leads to a substantial drop in their performance, \textit{highlighted in \crule[green!40]{.18cm}{.18cm}}. Several other classes also exhibit similar degradation. Our proposed approach ensures that previously similar classes retain their knowledge while effectively adapting to new ones by incorporating a controlled noise component.

A direct comparison between $\theta_{CTA}\ (w/o\ \mathcal{N})$ and $\theta_{Avg}\ (w\ \mathcal{N})$ reveals an important interaction. $\theta_{Avg}\ (w\ \mathcal{N})$ causes the cow class to collapse despite otherwise reasonable performance. This occurs because uniform noise applied to an unweighted average amplifies the representation variance of low-frequency classes, pushing their boundaries in conflicting directions. In contrast, $\theta_{CTA}\ (w/o\ \mathcal{N})$ fails specifically for classes with high within-task semantic similarity, where the absence of a noise buffer allows proximity-induced overlap (described in $\S$~\ref{sec: cbrs}) to corrupt the source class boundary. Together, these results demonstrate that neither component alone is sufficient. The attention module provides the correct semantic direction, while the noise provides the necessary geometric separation.\\

\noindent \textbf{Analysis on Stability-Plasticity Trade-off: }A fundamental challenge in CISS is navigating between retaining knowledge of previously learned classes (stability) and efficiently acquiring new ones (plasticity). We provide a granular analysis by independently tracking old-class mIoU (classes 1–15) and new-class mIoU (the cumulative number of new classes at each step) across all five incremental steps of the 15-1 setting, comparing SELECT against MBS, NeST, \input{Tables/wrap_trade}and BARM~\cite{DBLP:journals/corr/abs-2407-09838}. Results, visualized in Fig.~\ref{fig:tradeoff}, plot this as a stability-plasticity trade-off. SELECT (\crule[blue]{.18cm}{.18cm}) is superior over all baselines at every incremental step. At the final step, SELECT achieves 83.3\% old-class mIoU and 72.0\% new-class mIoU, outperforming MBS (82.3\%, 69.0\%), NeST (76.8\%, 54.4\%), and BARM (68.3\%, 27.2\%) on both dimensions simultaneously. SELECT occupies the upper-right at every step, confirming the method's efficiency.\\
Upon looking closely, we observe that NeST exhibits moderate plasticity but suffers significant stability degradation. Old-class mIoU declines by $\approx$6.0\% from base to final step, confirming our analysis in $\S$~\ref{sec:oursvsnest} that undifferentiated global distillation dilutes useful features with noise from irrelevant classes. BARM exhibits the most pronounced forgetting alongside the lowest plasticity ($\approx$27.2\% new-class mIoU at Step~5). MBS is the closest; its stability is comparable to SELECT, but its new-class mIoU consistently lags by $\approx$3\% at each step.\\

%% file: Tables/pascal_ade_final.tex
\begin{table*}[htb]
\centering
\caption{\small Performance comparison under different scenarios for \textit{overlapped} setting. $\ddagger$ implies results are reproduced from the official repository. $\dagger$ indicates results are excerpted from \cite{DBLP:journals/corr/abs-2407-11859, incrementer}. Best and second best results are marked in \crule[green!70]{.18cm}{.18cm} and \crule[green!15]{.18cm}{.18cm}, respectively. 
}
\label{tab:pascal_ade}
\vspace{10pt}
\setlength{\tabcolsep}{3pt}
\renewcommand{\arraystretch}{1}
\resizebox{\textwidth}{!}{
\centering
\begin{tabular}{lc ccc ccc ccc  ccc ccc ccc}
\toprule \midrule
\multirow{3}{*}{Methods} & \multirow{3}{*}{Backbone}  
& \multicolumn{9}{c}{\textbf{Pascal VOC}} & \multicolumn{9}{c}{\textbf{ADE20K}}  \\
\cmidrule(lr){3-11} \cmidrule(lr){12-20}
& & \multicolumn{3}{c}{19-1 (2 tasks)} & \multicolumn{3}{c}{15-5 (2 tasks)} 
& \multicolumn{3}{c}{15-1 (6 tasks)}  
& \multicolumn{3}{c}{100-50 (2 tasks)} & \multicolumn{3}{c}{50-50 (3 tasks)} 
& \multicolumn{3}{c}{100-10 (6 tasks)} \\
\cmidrule(lr){3-5} \cmidrule(lr){6-8} \cmidrule(lr){9-11} \cmidrule(lr){12-14}
\cmidrule(lr){15-17} \cmidrule(lr){18-20} 
& & 1-19 & 20 & All & 1-15 & 16-20 & All & 1-15 & 16-20 & All  
& 1-100 & 101-150 & All & 1-50 & 51-150 & All & 1-100 & 101-150 & All \\
\midrule
   SSUL \cite{DBLP:conf/nips/ChaKYM21} & Res101 &   77.8   &   49.8    &  76.5     & 78.4      & 55.8      & 73.0  & 78.4      & 49.0     & 71.4  & 42.8 & 17.5 & 34.5 & 49.1 & 20.1 & 29.8 & 42.9 & 17.7 & 34.5  \\
   SPPA \cite{DBLP:conf/eccv/LinWZ22} & Res101 & 76.5  & 36.2  & 74.6 & 78.1  & 52.9 &  72.1 & 66.2  & 23.3   & 56.0 & 42.9   & 19.9  & 35.2 & 49.8  & 23.9   & 32.5 & 41.0  & 12.5  & 31.5 \\
   RCIL$^{\dagger}$ \cite{DBLP:conf/cvpr/ZhangXLCC22}  & Res101 &   77.0   &   31.5    &  74.7   & 78.8      & 52.0      & 72.4    & 70.6      & 23.7     & 59.4    & 42.3      & 18.8      & 34.5     &  48.3     & 25.0      & 32.5     & 39.3      & 17.6      & 32.0     \\
   IDEC \cite{DBLP:journals/pami/ZhaoYS23}& Res101 & - & - & - & 78.0  & 51.8  & 71.8 & 77.0  & 36.5  & 67.3  & 42.0   & 18.2  & 34.1 & 47.4  & 26.0  & 33.1 & 40.3  & 17.6  & 32.7  \\
   LGKD+PLOP \cite{DBLP:conf/iccv/0002LLZWHMHL23} & Res101 & 76.5  & 42.9  & 75.7 & 78.7  & 56.1  & 73.9 & 69.3   & 30.9  & 61.1    & 43.6 &  25.7   & 37.5 & 49.4  & 29.4  & 36.0 & 42.1   & 22.0  & 35.4 \\
   STAR \cite{DBLP:conf/nips/ChenCLIK23} & Res101 & 78.0  & 47.1  & 76.5 & 79.5  & 58.9 &  74.6 & 79.5 &  50.6  & 72.6 & 42.4  & 24.2   & 36.4 & 48.7  & 27.2  & 34.4 & 42.0  & 20.6  & 34.9  \\
   REMINDER \cite{DBLP:conf/cvpr/PhanTPTB22} & Res101 & 76.5  & 32.3  & 74.4 & 76.1  & 50.7 & 70.1 & 68.3 & 27.2  & 58.5   & 41.6  & 19.2   & 34.1 & 47.1  & 20.4  & 29.4 & 39.0 & 21.3 & 33.1   \\
   BARM \cite{DBLP:journals/corr/abs-2407-09838} & Res101 & 78.2      & 42.2      & 76.4   & -      & -      & -    & 77.6      & 45.9      & 70.0   & 42.0      & 23.0      &  35.7    & 47.9      & 26.5      & 33.7     & 41.1      & 23.1      &  35.2   \\
   ADAPTER \cite{DBLP:journals/corr/abs-2412-12669} & Res101 & 78.0 &  50.7  & 76.7 & 79.7  & 59.7  & 75.0 & 79.9  & 51.9  &73.2  & 43.1  & 23.6  & 36.7 & 49.3   & 27.3  & 34.7 &42.9    & 19.9  & 35.3   \\
   MiB$^{\dagger}$ \cite{DBLP:conf/cvpr/CermelliMB0C20} &  ViT & 79.9      & 47.7      & 79.1    & 78.6      & 63.1      & 75.6    & 72.6      & 23.1      & 61.7   &  46.4     & 35.0      & 42.6     & 52.2      & 35.6      & 41.1    & 43.0      & 30.8      & 38.9      \\
   CoinSeg \cite{Zhang_2023_ICCV} & Swin-B  & 81.5  & 44.8  & 79.8 & 82.1  & 63.2    & 77.6 &82.7  & 52.5   & 75.5   &41.6   &26.7   & 36.6 & 49.0  & 28.9  & 35.6 & 42.1   & 24.5  & 36.2  \\
   INC$^{\dagger}$ \cite{incrementer} & ViT & \cellcolor{green!15}\textbf{82.5} & 61.0      & \cellcolor{green!70} \textbf{82.1}    & 82.5      & 69.2      & 79.9    & 79.6      & 59.6      & 75.6    & 49.4      & 35.6      & 44.8     & \cellcolor{green!15}\textbf{56.2}     & 37.8      & 43.9     & 48.5      & 34.6      & \cellcolor{green!15}\textbf{43.9}  \\
   MiB + NeST \cite{DBLP:conf/eccv/XieLXWZL24} & Swin-B & 79.7      & 60.0      & 78.8   & 81.2      & 67.4      & 77.9    & 77.0       & 53.3      & 71.4    &  42.8     & 27.8      & 37.9     & 49.7      & 29.3      & 36.2     & 41.8      & 23.8      & 35.9  \\ 
   PLOP + NeST \cite{DBLP:conf/eccv/XieLXWZL24} & Swin-B & 79.6      & \cellcolor{green!70}\textbf{70.2}  & 79.1   & 80.5      & 70.8      & 78.2 & 76.8      & 57.2      & 72.2   & 43.5      & 26.5      & 37.9     & 50.6      & 28.9      & 36.2     & 41.7      &  24.2     & 35.9   \\
   MBS$^{\ddagger}$ \cite{DBLP:journals/corr/abs-2407-11859} & ViT & 81.5    & 67.0      &   80.8   &  82.7     &  \cellcolor{green!15}\textbf{74.0}     &  80.5   &  82.3     & \cellcolor{green!15}\textbf{69.0}      &  79.0  &  47.7     & 35.6      & 43.7     & 54.4      & 37.1      &  42.9    & 47.7      & 31.5      &  42.3   \\ 
   CoGaMiD\cite{zhucontinual}  & Swin-B & - & -      & -  & -      & -      & -      & 83.2      & 61.2      & 78.0   &  43.9    &  27.3  &  38.4    &  49.9       &  29.8     &  36.6   &  43.7    &  26.5     &  38.0    \\ 
   EIR \cite{yin2025beyond} & Swin-B & - & -  & - & \cellcolor{green!15}\textbf{83.4}  & 68.6  &79.9 & \cellcolor{green!70}\textbf{83.6}  & 66.9  & \cellcolor{green!15}\textbf{79.6}   & 42.1 & 27.3  & 37.2 & 49.7 & 28.8  & 35.8 & 42.3 & 23.6  & 36.1 \\
   LBD \cite{wu2025learning} & ViT & 82.2      & \cellcolor{green!15}\textbf{70.0}  & \cellcolor{green!15}\textbf{81.6}      & 83.2      & 73.6      & \cellcolor{green!15}\textbf{80.8} & 81.9      & 66.6      & 78.1   & \cellcolor{green!15}\textbf{51.3}     & \cellcolor{green!15}\textbf{38.7}      & \cellcolor{green!15}\textbf{47.1}      & \cellcolor{green!15}\textbf{56.2}      & \cellcolor{green!15}\textbf{40.6}      & \cellcolor{green!15}\textbf{45.8}     & \cellcolor{green!15}\textbf{48.7}      & \cellcolor{green!15}\textbf{34.9}     & \cellcolor{green!70}\textbf{44.1}   \\   \midrule
   
   Ours & ViT & 
   \cellcolor{green!70}\textbf{83.0}  & \cellcolor{green!70}\textbf{70.2}   & \cellcolor{green!70}\textbf{82.1}   & \cellcolor{green!70}\textbf{83.9}   &  \cellcolor{green!70}\textbf{76.0}   &  \cellcolor{green!70}\textbf{81.6}   & \cellcolor{green!15}\textbf{83.3}   & \cellcolor{green!70}\textbf{72.0}   &  \cellcolor{green!70}\textbf{80.5}   &  \cellcolor{green!70}\textbf{54.1}  & \cellcolor{green!70}\textbf{40.3}   &  \cellcolor{green!70}\textbf{48.2}   &  \cellcolor{green!70}\textbf{56.4}   & \cellcolor{green!70}\textbf{42.3}   & \cellcolor{green!70}\textbf{47.1}   & \cellcolor{green!70}\textbf{50.4}  & \cellcolor{green!70}\textbf{35.7}   & \cellcolor{green!70}\textbf{44.1} \\ \midrule \bottomrule  
\end{tabular}
}
\end{table*}

%% file: Tables/ablation_sim_dissim.tex
\begin{wrapfigure}{l}{0.4\linewidth}
\vspace{-13pt}
\centering
\captionof{table}{\small Analyzing the impact of similar $\mathcal{C}_s$ and dissimilar $\mathcal{C}_{ds}$ classes on Pascal VOC dataset. \textbf{Bold} denotes the best result.}
\label{tab:ablation_simivsdissimi}
\vspace{5pt}
\setlength{\tabcolsep}{5pt}
\renewcommand{\arraystretch}{1}
\resizebox{\linewidth}{!}{
\begin{tabular}{ccccc ccc}
\toprule \midrule
\multirow{2}{*}{$\mathcal{C}_{s}$} & 
\multirow{2}{*}{$\mathcal{C}_{ds}$} & 
\multicolumn{3}{c}{15-1 (6 tasks)} & 
\multicolumn{3}{c}{15-5 (2 tasks)} \\
\cmidrule(lr){3-5} \cmidrule(lr){6-8}
& & 1-15 & 16-20 & All & 1-15 & 16-20 & All \\
\midrule
\cross & \checkmark & 78.5 & 36.6 & 71.5 & 79.6 & 66.8 & 76.4 \\
\checkmark & \cross & \textbf{83.3} & \textbf{72.0} & \textbf{80.5} & \textbf{83.9} & \textbf{76.0} & \textbf{81.6} \\
\midrule
\bottomrule
\end{tabular}
}
\end{wrapfigure}

%% file: Tables/ablation_similarity_metrics.tex
\begin{wrapfigure}{l}{0.4\linewidth}
\vspace{-13pt}
\centering
\captionof{table}{\small Ablation study on Pascal VOC. Analysis on the different similarity metrics \textbf{Bold} denotes the best result.}
\label{tab:ablation_3}
\vspace{5pt}
\setlength{\tabcolsep}{3pt}
\renewcommand{\arraystretch}{1}
\resizebox{.42\textwidth}{!}{
\centering
\begin{tabular}{cccc ccc}
\toprule \midrule
\multirow{2}{*}{Metrics} & 
\multicolumn{3}{c}{15-1 (6 tasks)} & 
\multicolumn{3}{c}{15-5 (2 tasks)} \\
\cmidrule(lr){2-4} \cmidrule(lr){5-7}
 & 1-15 & 16-20 & All & 1-15 & 16-20 & All \\
\midrule
Cosine   & 80.8  &  60.4  & 76.5 & 83.4 & 72.8  & 81.4    \\
Manhattan    & 78.5   &  44.9  & 71.2 & 83.2 & 73.7  & 81.4  \\
MSE    & 70.9   &  66.4  & 70.3 & 83.6 & 72.3  & 81.4     \\ \midrule
\textbf{Euclidean}  &   \textbf{83.3}       &  \textbf{72.0}     & \textbf{80.5}  & \textbf{83.9}      & \textbf{76.0}      & \textbf{81.6}    \\ \midrule
\bottomrule
\end{tabular}
}
\end{wrapfigure}

%% file: Tables/ablation_init_strat.tex
\begin{wrapfigure}{l}{0.4\linewidth}
\vspace{-10pt}
\centering
\captionof{table}{\small Analyzing the initialization strategy and loss configuration on Pascal VOC. \textbf{Bold} denotes the best result.}
\label{tab:ablation_init}
\vspace{5pt}
\setlength{\tabcolsep}{3pt}
\renewcommand{\arraystretch}{1}
\resizebox{.42\textwidth}{!}{
\centering
\begin{tabular}{llcc}
\toprule \midrule
Init Strategy & Loss Configuration & 15-1 (All) & 15-5 (All)\\ \midrule
$\theta_{bg}$ (Background)&$\mathcal{L}_{ce}+\mathcal{L}_{fd}+\mathcal{L}_{kd}$ & 67.4 & 78.9\\
$\theta_{CTA}$ (Ours)&$\mathcal{L}_{ce}+\mathcal{L}_{fd}+\mathcal{L}_{kd}$ & 76.1  & 80.0 \\
$\theta_{CTA}$ (Ours)&$\mathcal{L}_{ce}+\mathcal{L}_{fd}+\mathcal{L}_{kd}+\mathcal{L}_{ct}$ & \textbf{80.5} & \textbf{81.6} \\
\midrule
\bottomrule
\end{tabular}
}
\vspace{-10pt}
\end{wrapfigure}

%% file: Tables/ablation_loss.tex
\begin{wrapfigure}{l}{0.4\linewidth}
\vspace{-5pt}
\centering
\captionof{table}{\small Analysis on Pascal VOC dataset about the effectiveness of each loss function. \textbf{Bold} denotes the best result.}
\label{tab:ablation_2}
\vspace{5pt}
\setlength{\tabcolsep}{3pt}
\renewcommand{\arraystretch}{1}
\resizebox{.42\textwidth}{!}{
\centering
\begin{tabular}{lccc ccc ccc}
\toprule \midrule
\multirow{2}{*}{$\mathcal{L}_{ce}$} & \multirow{2}{*}{$\mathcal{L}_{fd}$} & \multirow{2}{*}{$\mathcal{L}_{kd}$} & \multirow{2}{*}{$\mathcal{L}_{ct}$} 
& \multicolumn{3}{c}{15-1 (6 tasks)} & \multicolumn{3}{c}{15-5 (2 tasks)} \\
\cmidrule(lr){5-7} \cmidrule(lr){8-10}
& & & & 1-15 & 16-20 & All & 1-15 & 16-20 & All \\
\midrule
\checkmark & \cross & \checkmark & \checkmark & 48.5 & 14.2 & 42.2 & 83.0 & 70.9 & 80.6 \\
\checkmark & \checkmark & \cross & \checkmark & 63.0 & 32.6 & 60.0 & 74.3 & 57.2 & 71.1 \\
\checkmark & \checkmark & \checkmark & \cross & 80.1 & 60.7 & 76.1 & 83.2 & 70.2 & 80.0 \\
\checkmark & \cross & \cross & \cross & 57.1 & 8.7 & 52.2 & 77.1 & 60.1 & 73.8 \\
\checkmark & \cross & \cross & \checkmark & 53.2 & 10.5 & 55.4 & 75.4 & 58.8 & 72.3 \\
\checkmark & \checkmark & \cross & \cross & 61.9 & 17.5 & 61.6 & 72.2 & 53.7 & 68.8 \\
\checkmark & \cross & \checkmark & \cross & 79.3 & 24.4 & 66.8 & \textbf{84.1} & 73.7 & \textbf{82.1} \\ \midrule
\checkmark & \checkmark & \checkmark & \checkmark & \textbf{83.3} & \textbf{72.0} & \textbf{80.5} & 83.9 & \textbf{76.0} & 81.6 \\ \midrule
\bottomrule
\end{tabular}
}
\vspace{-15pt}
\end{wrapfigure}

%% file: Tables/ablations_four.tex
\definecolor{sheep}{HTML}{B85450}
\definecolor{train}{HTML}{B85450}
\definecolor{aero}{HTML}{B85450}
\definecolor{boat}{HTML}{a484f5}
\definecolor{sofa}{HTML}{B85450}
\begin{table*}[h]
\centering
\caption{\small Effectiveness of knowledge transfer on Pascal VOC 15-1 setting. $\theta_{Best}$: token of the most similar class, $\theta_{Avg}$: average token of all similar classes, and $\theta_{CTA}$: adaptive token using Context Transfer Attention. \textbf{Bold} represents the best results. \crule[green!40]{.18cm}{.18cm} highlights the performance drops in similar classes.}
\label{tab:ablation_1}
\vspace{5pt}
\setlength{\tabcolsep}{2pt}
\renewcommand{\arraystretch}{1}
\resizebox{\linewidth}{!}{
      \centering
\begin{tabular}{l>{\columncolor{green!20}}ccc>{\columncolor{green!20}}cccccc>{\columncolor{green!20}}cccccccc>{\columncolor{green!20}}c>{\columncolor{green!20}}c>{\columncolor{green!20}}ccc}
\toprule
\midrule
\multirow{2}{*}{Methods} & \multicolumn{16}{c}{1-15} & \multicolumn{6}{c}{16-20} \\
\cmidrule(lr){2-17} \cmidrule(lr){18-23}
& aero & bike & bird & boat & bottle & bus & car & cat & chair & cow & table & dog & horse & motor & person & \textbf{Avg} 
& plant & sheep & sofa & train & tv & \textbf{Avg} \\
\midrule
$\theta_{Best}$ 
& 0.0 & 44.7 & 92.6 & 19.7 & 87.1 & 91.5 & 87.4 & 93.0 & 46.2 & 53.3 & 57.3 & 90.3 & 91.1 & 91.2 & 89.0 & 69.0 
& 39.7 & 55.8 & 40.1 & 55.3 & 57.6 & 49.7 \\

$\theta_{Avg}\ (w/o\ \mathcal{N})$ 
& 0.0 & 44.3 & 91.9 & 76.1 & 86.3 & 93.6 & 87.9 & 95.5 & 50.4 & 71.5 & 62.3 & 92.3 & 90.0 & 89.4 & 88.9 & 74.7 
& 56.1 & 0.1 & 36.6 & 43.4 & 62.6 & 39.8 \\

$\theta_{Avg}\ (w\ \mathcal{N})$ 
& 89.6 & 45.3 & 93.1 & 77.1 & 85.4 & 90.9 & 87.6 & 95.3 & 45.9 & 4.4 & 56.7 & 92.5 & 84.6 & 89.2 & 88.6 & 75.1 
& 70.0 & 0.0 & 30.9 & 44.5 & 59.6 & 41.0 \\

$\theta_{CTA}\ (w/o\ \mathcal{N})$ 
& 16.0 & 44.7 & 87.6 & 80.1 & 78.9 & 92.4 & 88.5 & 94.4 & 50.2 & 48.5 & 63.2 & 92.5 & 90.7 & 89.3 & 88.7 & 73.7 
& 72.1 & 9.3 & 44.5 & 42.1 & 57.7 & 45.1 \\

$\theta_{CTA}$ (Ours) 
& 93.7 & 49.1 & 90.3 & 81.4 & 87.8 & 95.2 & 92.0 & 95.3 & 51.9 & 83.1 & 64.1 & 94.2 & 91.3 & 90.7 & 89.4 & \textbf{83.3} 
& 68.2 & 89.0 & 48.6 & 86.9 & 67.3 & \textbf{72.0} \\
\midrule
\bottomrule
\end{tabular}
    }
\end{table*}


%% file: Tables/wrap_trade.tex
\setlength{\columnsep}{8pt}
\begin{wrapfigure}{l}{115pt}
\centering
\vspace{-20pt}
\includegraphics[width=0.35\textwidth]{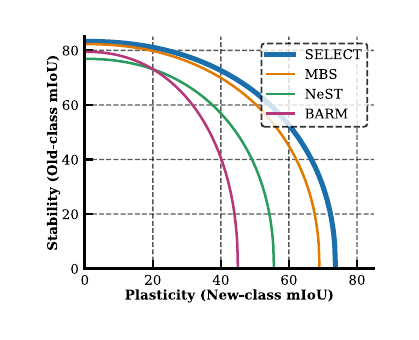}
\vspace{-10pt}
\caption{\small Comparative analysis of Stability-Plasticity trade-off.}
\label{fig:tradeoff}
\vspace{-10pt}
\end{wrapfigure}

%% file: sections/5_conclusion.tex
\section*{Conclusion}
In this work, we introduced {SELECT}, a novel framework that addresses an adaptive knowledge transfer strategy. By identifying semantic similarities between old and new classes and employing a selective transfer mechanism, SELECT provides a more principled approach to knowledge utilization. Our proposed loss function further strengthens the model by encouraging distinctiveness between transferred knowledge and the original representations of influential old classes. Finally, our empirical results demonstrate the effectiveness of our proposed approach.

\section*{Acknowledgement}
This work was supported in part by the Infosys Centre for Artificial Intelligence, IIIT-Delhi, and in part by the Institute Fellowship of the IIIT-Delhi.

%% file: sections/supplementary.tex
\appendix

\section{Complete Training Pipeline}
\label{sec:pipeline}

In this section, we present our complete training pipeline which represents identifying knowledge from previously learned similar classes and selectively transferring that knowledge to initialize the new class. 

\input{Algorithm/strategy.tex}

\section{Additional Implementation Details}
We perform all experiments using the PyTorch framework (version 1.10.1) on a single workstation with an NVIDIA A100 GPU. We ensure that all experiments are performed under fair, consistent conditions in a unified environment. As mentioned in Table 2 in the paper, we reimplement the results of MBS \cite{DBLP:journals/corr/abs-2407-11859} for fair and direct comparison. Similarly, for additional analysis on NeST \cite{DBLP:conf/eccv/XieLXWZL24}, REMINDER~\cite{DBLP:conf/cvpr/PhanTPTB22}, and BARM \cite{DBLP:journals/corr/abs-2407-09838}, we reimplement their results.

We obtained the results by running their respective official, publicly available source codes. We used their implementation details without modifying the core architecture or loss functions. To ensure a fair comparison, our proposed method (SELECT) and MBS were trained and evaluated in the same unified environment. We will publicly release the source code for our proposed method to facilitate better reproduction.
\section{Analysis on Class Similarity}
\subsection{Image-wise Disparity}
\begin{figure*}[h]
    \centering
    \includegraphics[width=\textwidth]{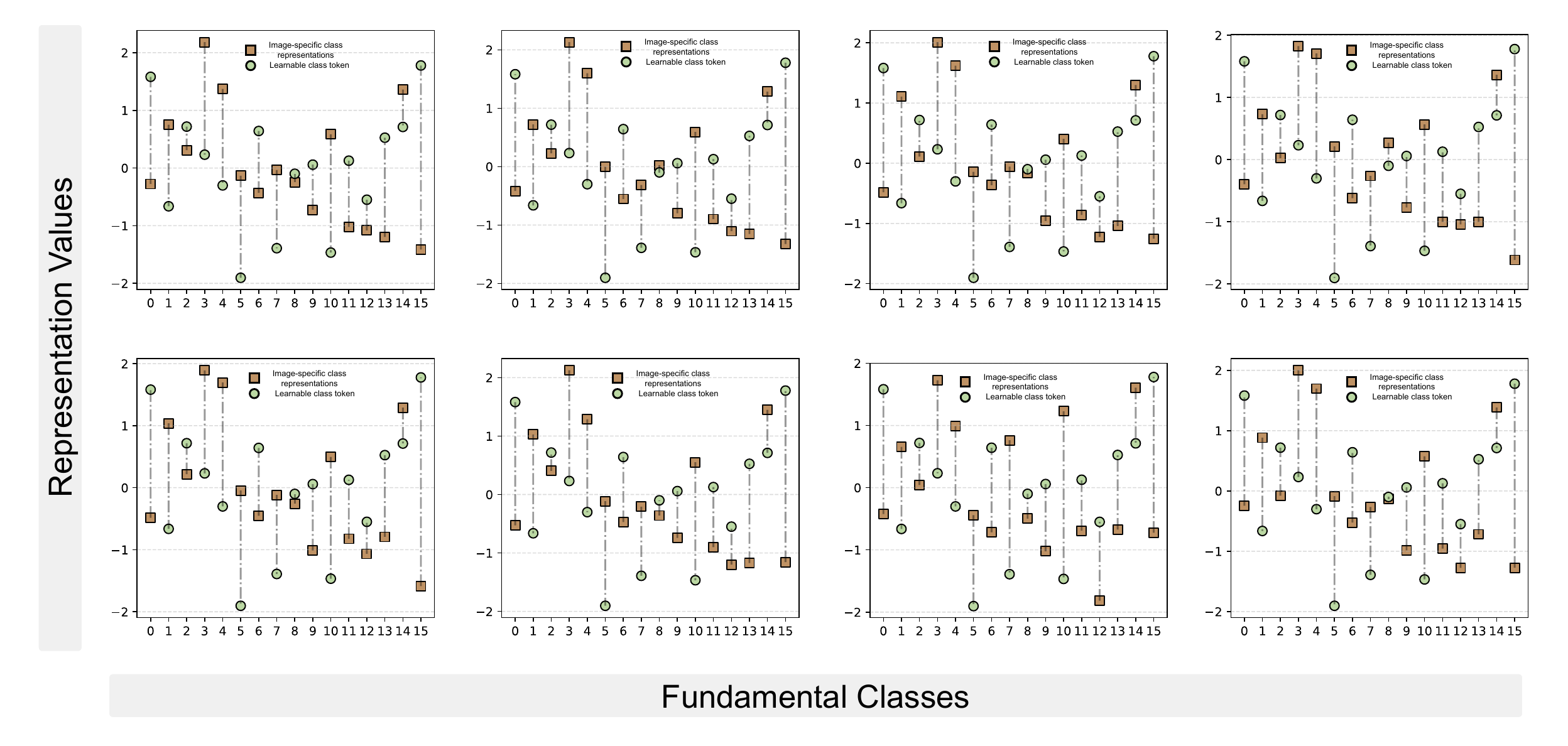}
    \vspace{5pt}
    \caption{\small Representational perturbation between $e_{\mathcal{C}_{0:t-1}}$ and $\hat{e}_{\mathcal{C}_{(0:t-1)}}$ for randomly sampled images of ``sheep''.}
    \label{fig:analysis8}
\end{figure*}

In the main paper, we demonstrated the strong semantic relevance and lower average perturbation across new classes. In Fig.~\ref{fig:analysis8}, we randomly sample a few images (from the current task's dataset) and visualize these perturbations.

It is evident that all images exhibit deviation within their respective classes, with some classes showing minimal deviation, indicating strong similarity in their characteristics. This similarity is utilized to identify the most similar class.

\subsection{Further Analysis on Semantic Similarity}

To further analyze the effect of initialization, we conduct an additional controlled study across eight experimental configurations. In each study, three base classes are trained, after which ``Sheep'' is introduced as the incremental class. For each study, we compare two conditions: a similar model, in which the incremental class (sheep) token is initialized from the base classes deemed most semantically similar, and a dissimilar model, in which the incremental class (sheep) token is initialized from semantically unrelated base classes.

\begin{figure*}[ht]
    \centering
    \includegraphics[width=\textwidth]{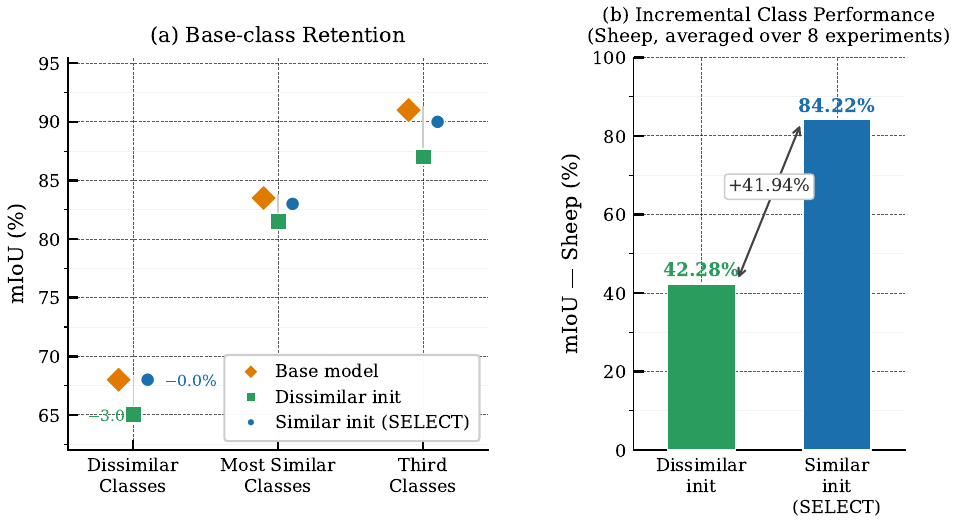}
    \caption{\small a) Scatter plot for analyzing the choice of initialization source over eight experiments; b) Incremental class performance averaged over eight experiments.}
    \label{fig:sirs}
\end{figure*}

Fig.~\ref{fig:sirs}(b) shows the incremental class performance averaged over the 8 configurations. The similar model achieves 84.22\% mIoU on Sheep, while the dissimilar model achieves only 42.28\%. When the model begins from a semantically similar point, a source class that shares visual features with the target class, the decoder can adapt with far fewer gradient steps.
Fig.~\ref{fig:sirs}(a) further reveals that the choice of initialization source also affects how well the base classes are retained. For all class groups, the similar model consistently outperforms the dissimilar model. The performance gap is most pronounced in the third class group, indicating that catastrophic forgetting propagates beyond the directly interfering classes. A structurally poor initialization forces the optimizer to make larger weight updates across the decoder, disturbing representations that were not directly involved. 

\begin{figure*}[ht]
    \centering
    \includegraphics[width=\textwidth]{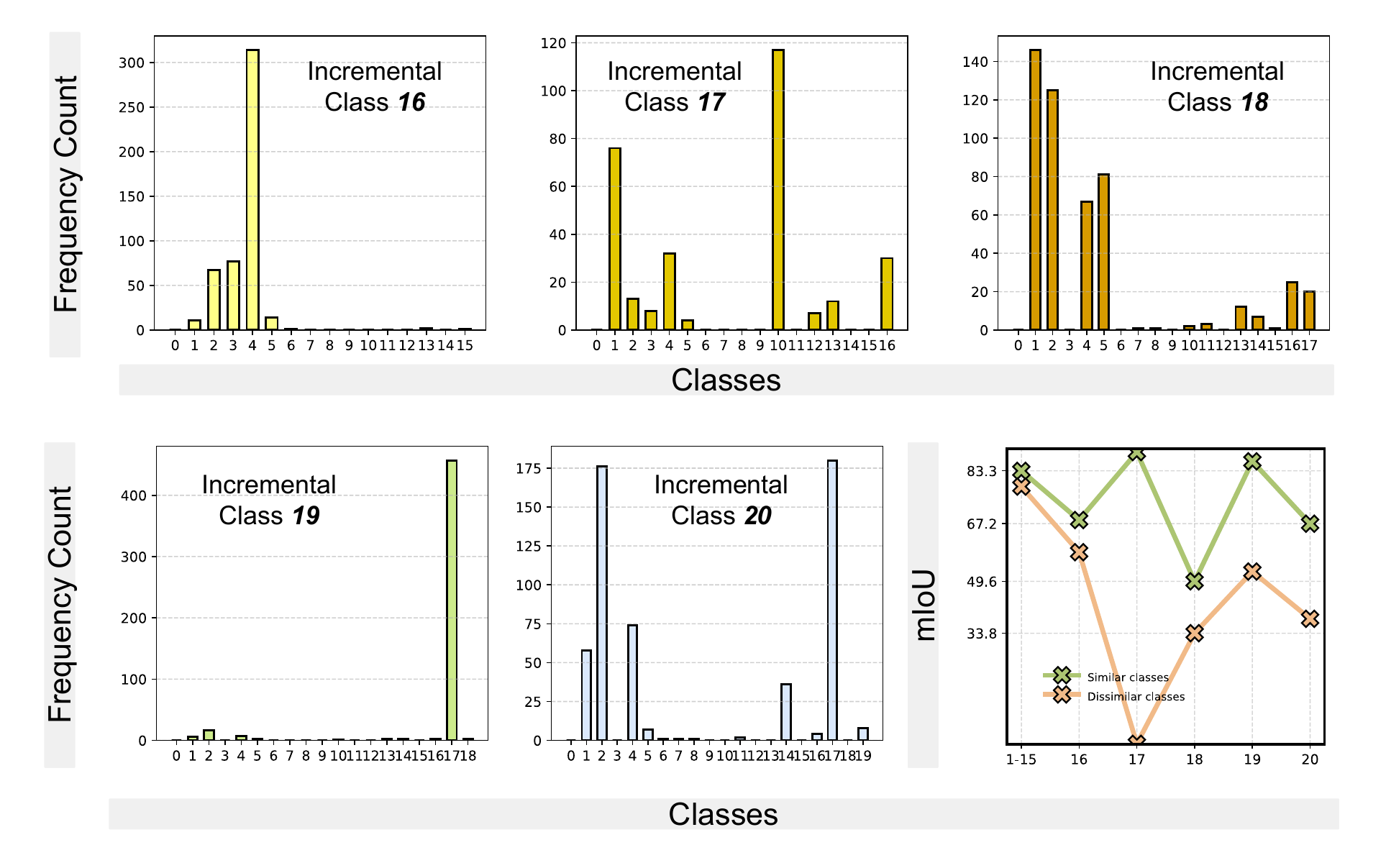}
    \caption{\small Visualization of class distribution across the incremental scenario in the 15-1 setting. Additionally, there is a visual comparison of similar and dissimilar classes.}
    \label{fig:analysis9}
\end{figure*}

Finally, our proposed approach leverages the characteristics of a well-trained representational space, where features are not random but semantically clustered. In Fig.~\ref{fig:analysis9}, we provide the visualization of class distribution for all incremental scenarios in the 15-1 setting. From the figure, it is evident that there is a selective number of classes in each incremental class, which resembles the distribution. Furthermore, we provide an ablation study comparing performance between similar and dissimilar classes.

\subsubsection{Similarity Detection Cost}
Similarity detection is a single forward pass of current-task data through the frozen old model; the cost of computing Euclidean distances is negligible. We report the average processing speed (imgs/sec per task) for ADE20K and compare with NeST (ours/NeST), 100-50: 70.0/10.2, $\|$ 50-50: 69.4/10, $\|$ 100-10: 71.6/11.1, $\|$ 100-5: 74.0/13.

\input{Tables/pascal_tot}
\input{Tables/pascal_dis.tex}
\section{Additional Analysis and Results}
\subsection{Diverse Challenging Scenarios}
In the paper, we present a unified set of experimental results on Pascal VOC and ADE20K in an overlapped scenario across selected settings. Here, we perform additional experiments on the aforementioned datasets.

In particular, Table~\ref{tab:pascal_supp} extends the quantitative results by comparing with recent works and providing results for the additional 5-3 (6 tasks) setting. 
In the 5-3 setting, SELECT scores 78.6\% in the ``All'' setting mIoU, compared to MBS's 77.0\%. The 5-3 setting presents two compounding challenges specific to our approach. First, with only 5 base classes, the semantic space is sparser. Each incremental class has fewer potential similar classes, reducing the informativeness of $\mathcal{C}_s$. Second, with only three classes per step, the per-class sample count during similarity detection is low, making the Euclidean deviation metric noisier. 
This analysis directly motivates adaptive thresholding as a future direction. 

Additionally, we perform experiments (demonstrated in Table~\ref{tab:pascal-dis}) in the disjoint scenario for 19-1 (2 tasks), 15-5 (2 tasks), and 15-1 (6 tasks) settings. Our proposed approach provides a strong competitive advantage over all previous approaches while demonstrating effectiveness.

In Table~\ref{tab:pascal_ade}, we demonstrate detailed experimental comparison for ADE20K dataset in the overlapped scenario. As mentioned in the paper, ADE20K presents more diverse and robust scenes, making it a challenging benchmark. Upon comparison with prior work, our proposed approach shows strong performance and surpasses it across almost all settings.

\input{Tables/ade_complete}

\subsection{Comparative Analysis over class-wise performance}

\begin{figure*}
    \centering
    \includegraphics[width=\linewidth]{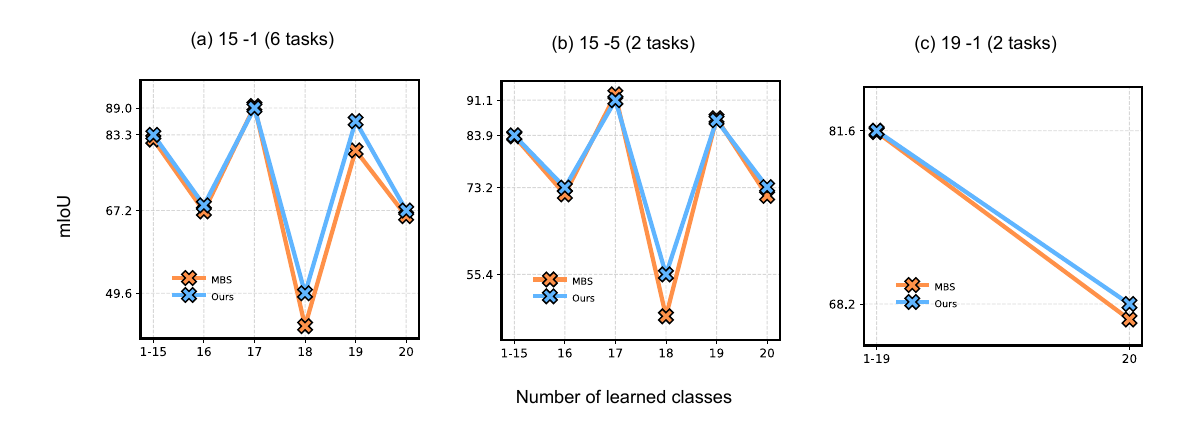}
    \caption{\small Incremental class-wise performance across different settings for Pascal VOC dataset.}
    \label{fig:analysis_pascal}
\end{figure*}

In Fig.~\ref{fig:analysis_pascal}, we visualize the performance of individual incremental classes for three common settings (15-1, 15-5, and 19-1) and compare the performance of our proposed method with MBS~\cite{DBLP:journals/corr/abs-2407-11859}. From the figure, we observe that both methods achieve similar performance in the fundamental classes across all settings. However, in the incremental scenario, there is a performance gap between the two approaches. Specifically, we see a performance drop in~\cite{DBLP:journals/corr/abs-2407-11859} across all settings.

\subsection{Additional Qualitative Results}
\begin{figure*}[h]
    \centering
    \includegraphics[width=\textwidth]{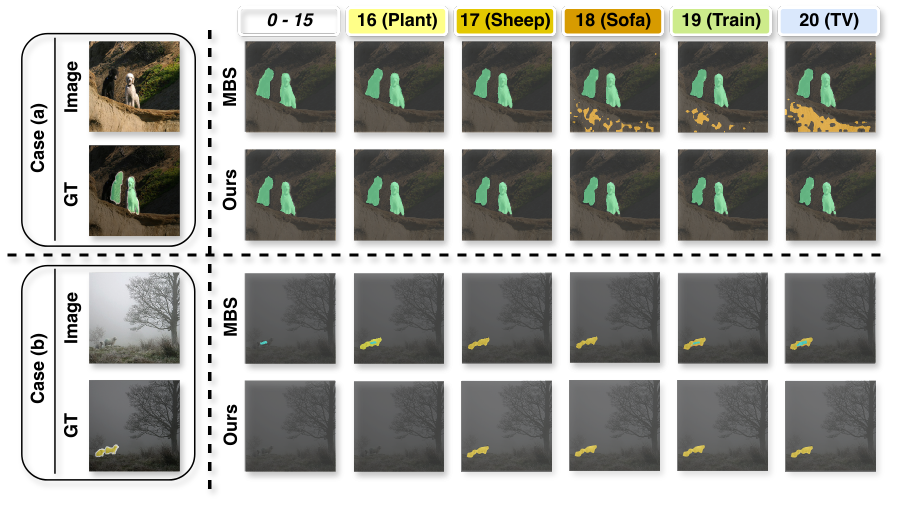}
    \caption{\small Additional visual comparison on the 15-1 setting of the Pascal VOC between MBS \cite{DBLP:journals/corr/abs-2407-11859} and ours.}
    \label{fig:pascal_vis_1}
\end{figure*}

\begin{figure*}[!h]
    \centering
    \includegraphics[width=\textwidth]{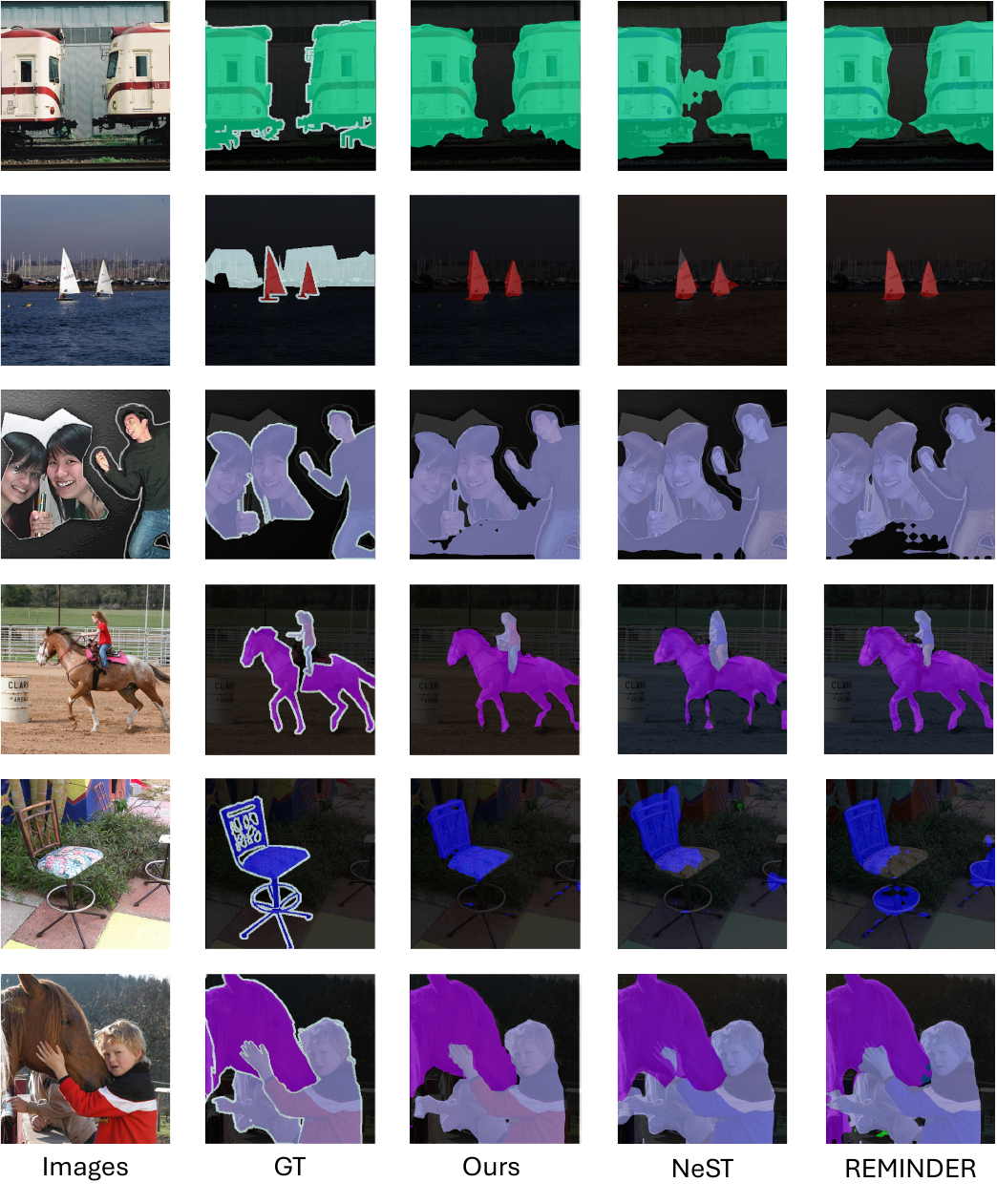}
    \caption{\small Visual comparison on the final segmentation prediction in 15-5 setting of the Pascal VOC among SELECT (Ours), REMINDER~\cite{DBLP:conf/cvpr/PhanTPTB22}, and NeST~\cite{DBLP:conf/eccv/XieLXWZL24}.}
    \label{fig:pascal_vis_2}
\end{figure*}

In this section, we present more qualitative comparisons with recent state-of-the-art approaches in Figs.~\ref{fig:pascal_vis_1}-\ref{fig:pascal_vis_2}.

In Fig.~\ref{fig:pascal_vis_1}, we provide the qualitative results of our proposed approach compared with~\cite{DBLP:journals/corr/abs-2407-11859}. We observe that~\cite{DBLP:journals/corr/abs-2407-11859} partially retains knowledge of previously learned objects; however, as new incremental tasks arise, it starts producing many false positives. As the task progresses, these false positives also get denser. On the other hand, our proposed approach maintains learning from previous classes while adapting to new knowledge more effectively. For example, in case(b), the sheep (in our case) is learned in the third task and is maintained thereafter. Conversely,~\cite{DBLP:journals/corr/abs-2407-11859} maintains part of the knowledge learned from the fundamental task and carries it forward.

Similarly, in Fig.~\ref{fig:pascal_vis_2}, we provide additional qualitative results of our proposed approach compared with~\cite{DBLP:conf/cvpr/PhanTPTB22, DBLP:conf/eccv/XieLXWZL24}. In this figure, we present the final segmentation results in a PascalVOC 15-5 setting. We observe that our proposed approach has close pixel-wise similarity with the ground truth as compared to others, demonstrating its robustness.

\subsection{SELECT on CNN backbone}
Based on previous work, to ensure a fair comparison, we provide an analysis that generalizes our proposed selective context transfer strategy using a CNN-based backbone. To validate this, we integrated our method into two standard CNN-based baselines (with Res101 backbones): MiB and PLOP and compared them against NeST. As shown in Table~\ref{tab:cnn}, our method consistently outperforms on these CNN baselines. In particular, on the 15-1 setting, our strategy improves over standard MiB by $\approx$20\% across all classes and over PLOP by $\approx$15\%, demonstrating robustness with a CNN-based backbone.

\input{Tables/pascal_cnn}
\subsection{SELECT \textit{vs} NeST}
In the main paper, we originally reported the results for NeST~\cite{DBLP:conf/eccv/XieLXWZL24} with the Swin-B backbone. In Table~\ref{tab:nest_vit}, we conduct a fair evaluation with the ViT-B backbone as well. The results presented in the table are averaged across all the classes for the 15-1, 15-5, and 10-1 settings. These benchmark results are extracted from NeST itself. From the table, we observe that our proposed approach significantly outperforms NeST in all tasks, demonstrating its effectiveness.
\input{Tables/nest_vit}

\begin{figure*}[ht]
    \centering
    \includegraphics[width=\linewidth]{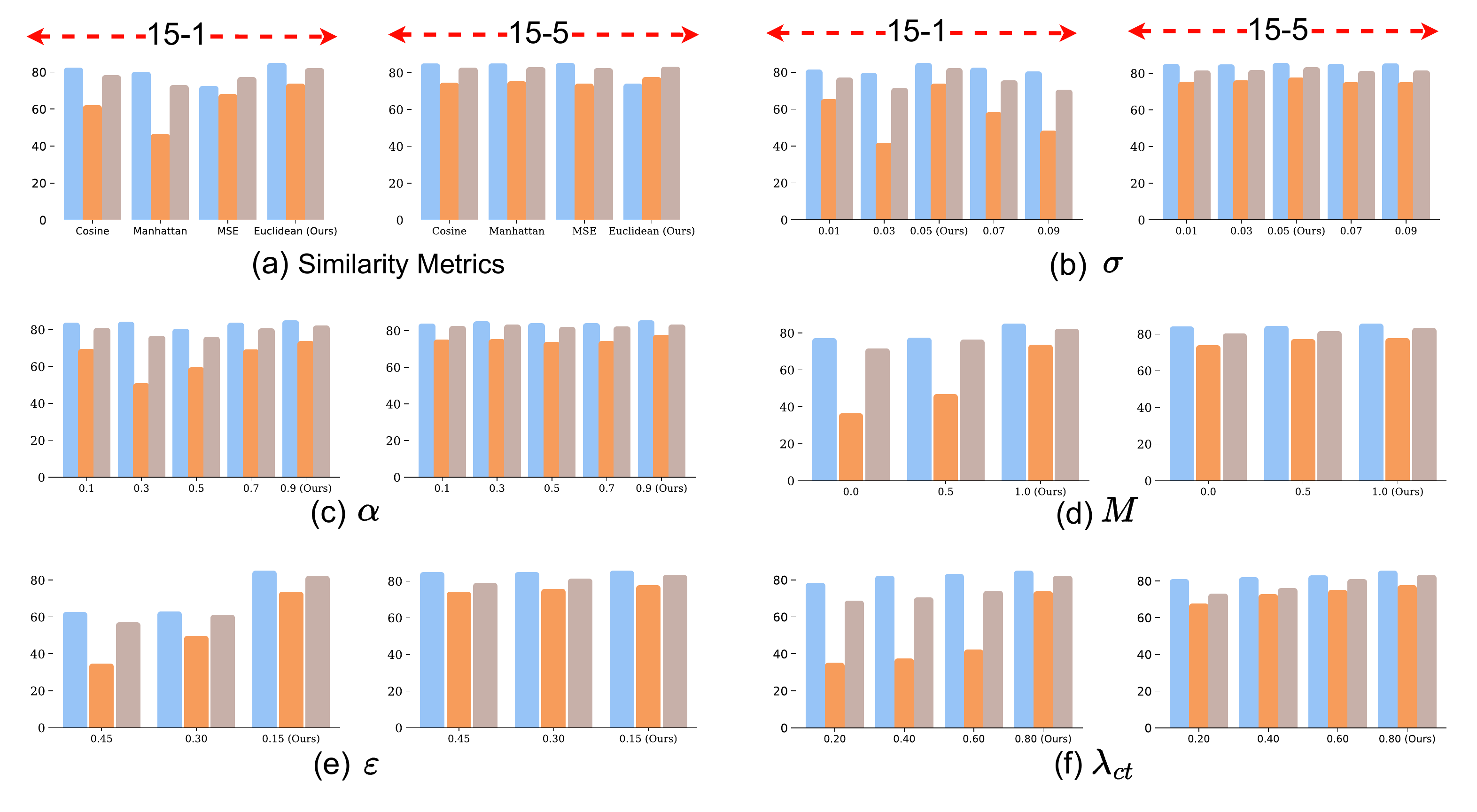}
    \caption{\small Analyzing different hyperparameters on PASCAL VOC 15-1 and 15-5 setting. (a) Different Similarity Metrics; (b) standard deviation $\sigma$; (c) noise trade-off $\alpha$; (d) Context transfer loss Margin $M$; (e) threshold frequency $\varepsilon$; (f) Context transfer loss weight $\lambda_{ct}$).}
    \label{fig:fin_abl}
\end{figure*}

\subsection{Additional Analysis over different hyperparameters}
In the paper, we provide a preliminary study on the hyperparameter. Below, we present the detailed analysis of all the hyperparameters used in our proposed approach.
\subsubsection{Different Similarity Metrics. }As mentioned in the paper, the similarity metric identifies the similarity between learned class tokens and image-specific class representations. In the paper, we presented the visual analysis in a 15-1 setting. In Fig.~\ref{fig:fin_abl}(a), we present the analysis in both 15-1 [left] and 15-5 [right] settings. We observe that the Euclidean metric performs significantly better than all other metrics, focusing on both magnitude and orientation.

\subsubsection{Standard Deviation $\sigma$. }We perform this study to determine the most appropriate $\sigma$ in the Gaussian noise $\mathcal{N}$ in $\S$3.3.2. Fig.~\ref{fig:fin_abl}(b) presents our findings. Standard deviation ($\sigma$) of the Gaussian noise plays a crucial role during knowledge transfer. We conduct an ablation study and found that $\sigma$=0.05 consistently yields the best performance across both tasks. This value represents an effective trade-off. Smaller values (\textit{e.g.} 0.01) fail to diversify the initialization meaningfully, while larger values (\textit{e.g.} 0.10) inject excessive noise.

\subsubsection{Noise Weight $\alpha$. }In $\S$3.3.2, we discuss about the Context Transfer Attention (CTA), and we add a controlled noise component $\mathcal{N}$ to provide a distinction between previously learned similar classes $\mathcal{C}_s$ and the new class $\mathcal{C}_t$. To maintain the balance between $\mathcal{C}_s$ and $\mathcal{N}$, we include a weighting hyperparameter $\alpha$. In Fig.~\ref{fig:fin_abl}(c), we analyze this hyperparameter by providing different values. From the figure, it is evident that our proposed strategy outperforms the alternative values for the $\alpha$. Particularly, when $\alpha$ is low (\textit{e.g.} 0.1), the representation is more influenced by the noise component, introducing flexibility. Conversely, when $\alpha$ (\textit{e.g.} 0.9) is high, the representation remains close to the original, offering more guidance. In the intermediate range, the representation shows the ambiguous behaviour that is neither sufficiently guided nor flexible, which explains the observed performance dip. 


\subsubsection{Context Transfer Loss Margin $M$. }The $\mathcal{L}_{ct}$ margin $M$ (demonstrated in  Fig.~\ref{fig:fin_abl}(d)) dictates how strictly the model forces separation between the new class and similar old classes. At $M=0.0$, we observe the lowest performance, as it fails to enforce a distinct decision boundary. While at $M=1.0$, we observe the highest performance. A larger margin provides a stronger constraint, compelling the model to push the new class representation sufficiently far away from prior similar classes. Hence, 1.0 is the appropriate choice.

\subsubsection{Threshold Frequency $\varepsilon$. } For the frequency threshold, our analysis (demonstrated in  Fig.~\ref{fig:fin_abl}(e)) consistently shows that 0.15 is the optimal value. When we increased the threshold to 0.30 or 0.45, the performance declined. Including a larger pool of similar classes (top 45\%) introduces semantic dilution. A tighter threshold (0.15) ensures that we only transfer knowledge from highly relevant predecessors, preventing confusion for the new class.
\subsubsection{Context Transfer Loss Weight $\lambda_{ct}$. }$\lambda_{ct}$ provides the weight to the proposed Context Transfer Loss ($\mathcal{L}_{ct}$). As visualized in Fig.~\ref{fig:fin_abl}(f), $\lambda_{ct}$ performs significantly better at 0.8 for both 15-1 and 15-5 settings. This weight value, in addition to $M$, significantly contributes in protecting the boundaries of the old classes while learning the new ones.

%% file: Algorithm/strategy.tex
\begin{algorithm}
\caption{Training Strategy for the proposed SELECT.}
\label{algo:strategy}
\begin{algorithmic}[1]
\Require Training dataset $\mathcal{D}_t = \{(x_i, y_i)\}_t$ for current task $t$, model with previous task's weights $f_{t-1}$, current task's model $f_t$, total tasks $T$.
\Ensure Initial class representation for new task $\theta_{CTA}$
\For{$t \in \{1,2,\dots,T\}$}
    \For{$(x_i, y_i) \in \mathcal{D}_t$} \Comment{Class similarity detection.}
        \State $\mathcal{C}_s \gets \textit{ClassSimilarity}(x_i, y_i, f_{t-1}$)
    \EndFor
    \State $\theta_s \gets \mathcal{C}_s$ \Comment{Corresponding tokens of previous similar classes.}
    \State $\hat{\theta}_{CTA} \gets \textit{Attention}(\theta_{s})$ \Comment{Context transfer attention.}
    \State $f_t \gets Update(f_t|\ \hat{\theta}_{CTA})$
    \While{not converged} \Comment{Training incremental setting.}
        \State train $f_t$
    \EndWhile
\EndFor
\end{algorithmic}
\end{algorithm}

%% file: Tables/pascal_tot.tex
\begin{table*}[ht]
\centering
\caption{\small Performance comparison on Pascal VOC under different scenarios for \textit{overlapped} setting. $\ddagger$ implies results are reproduced from the official repository. $\dagger$ indicates results are excerpted from \cite{DBLP:journals/corr/abs-2407-11859, incrementer}.}
\vspace{10pt}
\label{tab:pascal_supp}
\setlength{\tabcolsep}{4pt}
\renewcommand{\arraystretch}{1}
\resizebox{\linewidth}{!}{
\centering
\begin{tabular}{lc ccc ccc ccc ccc}
\toprule \midrule
\multirow{2}{*}{Methods} & \multirow{2}{*}{Backbone} & \multicolumn{3}{c}{19-1 (2 tasks)} & \multicolumn{3}{c}{15-5 (2 tasks)} & \multicolumn{3}{c}{15-1 (6 tasks)}   & \multicolumn{3}{c}{5-3 (6 tasks)}                        \\ 
\cmidrule(lr){3-5} \cmidrule(lr){6-8} \cmidrule(lr){9-11} \cmidrule(lr){12-14}
                 
        &   &  1-19     & 20      & All   &  1-15     & 16-20      & All    &  1-15     & 16-20      & All  &  1-5     & 6-20      & All  \\ \midrule
   MiB$^{\dagger}$ \cite{DBLP:conf/cvpr/CermelliMB0C20} & Res101 & 70.2      & 22.1      & 67.8      & 75.5      & 49.4      & 69.0    & 35.1      & 13.5      & 29.7   & 57.1      & 42.6      & 46.7 \\ 
   PLOP$^{\dagger}$ \cite{DBLP:conf/cvpr/DouillardCDC21} & Res101 & 75.4      & 37.4      & 73.5        & 75.7      & 51.7      & 70.1   &  65.1     & 21.1      & 54.6      & 17.5      & 19.2      & 18.7 \\  
   RCIL$^{\dagger}$ \cite{DBLP:conf/cvpr/ZhangXLCC22} & Res101 &   77.0   &   31.5    &  74.7   & 78.8      & 52.0      & 72.4    & 70.6      & 23.7     & 59.4       & -      & -      & - \\
   SSUL \cite{DBLP:conf/nips/ChaKYM21} & Res101 &   77.8   &   49.8    &  76.5        & 78.4      & 55.8      & 73.0   & 78.4      & 49.0     & 71.4   &71.3   & 53.2     &58.4  \\ 
   PLOP + Cs$^2$K \cite{DBLP:conf/eccv/CongCLS24} & Res101 &   -  &   -   &    -      &    -   &  -     &  -  &   77.9     &  46.4     & 70.4 & 58.4  &53.4     &54.8  \\ 
   MiB + NeST \cite{DBLP:conf/eccv/XieLXWZL24} & Res101 & 71.7      & 28.2      & 69.7     & 77.1      & 50.1      & 70.7    & 61.7       & 20.4      & 51.8       & -      & -      & - \\ 
   PLOP + NeST \cite{DBLP:conf/eccv/XieLXWZL24} & Res101 & 77.0      & 49.1      & 75.7    & 77.6      & 55.8      & 72.4        & 72.2      & 33.7      & 63.1       & -      & -      & - \\ 
   RCIL + NeST \cite{DBLP:conf/eccv/XieLXWZL24} & Res101 & 77.0      & 33.3      & 74.9  & 79.0      & 52.8      & 72.8  & 71.9      & 28.0      & 61.4       & -      & -      & - \\  
   BARM \cite{DBLP:journals/corr/abs-2407-09838} & Res101 & 78.2      & 42.2      & 76.4   & -      & -      & -   & 77.6      & 45.9      & 70.0   & 71.3      & 57.0      & 61.1 \\ 
   IDEC \cite{DBLP:journals/pami/ZhaoYS23} & Res101 & - & - & - & 78.0  & 51.8  & 71.8 & 77.0  & 36.5  & 67.3  & 67.1  & 49.0  & 54.1 \\
   STAR \cite{DBLP:conf/nips/ChenCLIK23} & Res101 & 78.0  & 47.1  & 76.5 & 79.5  & 58.9 &  74.6 & 79.5 &  50.6  & 72.6  & 71.9  & 61.5  & 64.4 \\
   ADAPTER \cite{DBLP:journals/corr/abs-2412-12669} & Res101 & 78.0 &  50.7  & 76.7 & 79.7  & 59.7  & 75.0 & 79.9  & 51.9  &73.2  & 73.8  & 61.9  & 65.3 \\
   EIR \cite{yin2025beyond} & Res101 & - & -  & - & 79.1  & 58.4  &74.2 & 79.4  & 52.6  &73.0  & 74.6  & 63.1  & 66.4 \\
   CoGaMiD\cite{zhucontinual}  & Res101 & - & -      & -  & -      & -      & -      & 80.1      & 53.6      & 73.8            & 73.7     & 63.1      & 66.1      \\
   CoinSeg \cite{Zhang_2023_ICCV} & Swin-B  & 81.5  & 44.8   & 79.8     & 82.1  & 63.2    & 77.6  & 82.7  & 52.5   & 75.5       & -    & -    & - \\ 
   MiB$^{\dagger}$ \cite{DBLP:conf/cvpr/CermelliMB0C20} & ViT  & 79.9      & 47.7      & 79.1    & 78.6      & 63.1      & 75.6   & 72.6      & 23.1      & 61.7    & 33.4      & 43.2      & 42.9 \\ 
    RBC$^{\dagger}$ \cite{DBLP:conf/eccv/ZhaoYFL22} & ViT & 80.2      & 38.8      & 79.0     & 78.9      & 62.0      & 75.5  & 75.9      & 40.2      & 68.2    & - & - & -  \\
   INC$^{\dagger}$ \cite{incrementer} & ViT & 82.5 & 61.0      & 82.1    & 82.5      & 69.2      & 79.9   & 79.6      & 59.6      & 75.6        & -      & -      & - \\ 
   MiB + NeST \cite{DBLP:conf/eccv/XieLXWZL24} & Swin-B & 79.7      & 60.0      & 78.8   & 81.2      & 67.4      & 77.9   & 77.0       & 53.3      & 71.4       & -      & -      & - \\ 
   PLOP + NeST \cite{DBLP:conf/eccv/XieLXWZL24} & Swin-B & 79.6      & 70.2  & 79.1   & 80.5      & 70.8      & 78.2 & 76.8      & 57.2      & 72.2       & -      & -      & - \\ 
CoGaMiD\cite{zhucontinual}  & Swin-B & - & -      & -  & -      & -      & -      & 83.2      & 61.2      & 78.0            & 79.9     & 72.7      & 74.7      \\ 
EIR \cite{yin2025beyond} & Swin-B & - & -  & - & 83.4  & 68.6  &79.9 & 83.6  & 66.9  &79.6  & 74.5  & 73.0  & 73.4 \\
MBS$^{\ddagger}$ \cite{DBLP:journals/corr/abs-2407-11859} & ViT & 81.5    & 67.0      &   80.8   &  82.7     &  74.0     &  80.5   &  82.3     & 69.0      &  79.0  & 76.7      & 77.3      & 77.0  \\ \midrule   
   Ours & ViT &  83.0    & 70.2  & 82.1    & 83.9     &  76.0     &  81.6   & 83.3  & 72.0   &  80.5    & 77.9      & 77.8      & 78.6 \\  \midrule \bottomrule  
\end{tabular}
}
\end{table*}

%% file: Tables/pascal_dis.tex
\begin{table*}[!h]
\centering
\caption{\small Additional performance comparison on Pascal VOC under different scenarios for \textit{disjoint} setting. $\ddagger$ implies results are reproduced from the official repository. $\dagger$ indicates results are excerpted from \cite{DBLP:journals/corr/abs-2407-11859, incrementer}.} 
\label{tab:pascal-dis}
\vspace{10pt}
\setlength{\tabcolsep}{3pt}
\renewcommand{\arraystretch}{1}
\resizebox{.8\linewidth}{!}{
\centering
\begin{tabular}{lcccccccccc}
\toprule \midrule
\multirow{2}{*}{Methods} & \multirow{2}{*}{Backbone} & \multicolumn{3}{c}{19-1 (2 tasks)}                        & \multicolumn{3}{c}{15-5 (2 tasks)}                        & \multicolumn{3}{c}{15-1 (6 tasks)}                        \\
\cmidrule(lr){3-5} \cmidrule(lr){6-8} \cmidrule(lr){9-11}
            &      &  1-19     & 20      & All   & 1-15      & 16-20      & All  & 1-15      & 16-20      & All \\ \midrule

   MiB$^{\dagger}$ \cite{DBLP:conf/cvpr/CermelliMB0C20} & Res101  & 69.6      & 25.6      & 67.4     & 71.8      & 43.3      & 64.7       & 46.2      & 12.9      & 37.9      \\ 
   PLOP$^{\dagger}$ \cite{DBLP:conf/cvpr/DouillardCDC21} & Res101  & 75.4      & 38.9      & 73.6     & 71.0      &  42.8     & 64.3      &  57.9     & 13.7      & 46.5   \\ 
   RCIL$^{\dagger}$ \cite{DBLP:conf/cvpr/ZhangXLCC22} & Res101  &  -     &  -  &   -     & 75.0      & 42.8      & 67.3     & 66.1      & 18.2      & 54.7   \\ 
   SPPA \cite{DBLP:conf/eccv/LinWZ22} & Res101  & 75.5  & 38.0  & 73.7 & 75.3 &  48.7  & 69.0  & 59.6  & 15.6  & 49.1  \\ 
   STAR \cite{DBLP:conf/nips/ChenCLIK23} & Res101  &  77.9   & 43.4   & 76.2 & 78.4   & 57.4  &  73.4 & 78.1  & 46.6  & 70.6 \\
    ADAPTER \cite{DBLP:journals/corr/abs-2412-12669} & Res101  & 78.0  &  46.1   & 76.5 & 78.9   & 58.2   & 73.9 & 78.6   & 49.0  & 71.5 \\
    CoGaMiD \cite{zhucontinual} & Res101 & 79.8 & 46.4      & 78.2  & 78.9      & 58.2      & 74.0      & 78.9 & 49.2      & 71.8  \\
   MiB$^{\dagger}$ \cite{DBLP:conf/cvpr/CermelliMB0C20} & ViT & 80.6      & 45.2      & 79.6    & 75.0      & 59.9      & 72.3     & 66.7      & 26.3      & 58.3    \\
   RBC$^{\dagger}$ \cite{DBLP:conf/eccv/ZhaoYFL22} & ViT & 80.9      & 42.1      & 79.7     & 77.7      & 59.1      & 74.0       & 69.0      & 28.4      & 60.5   \\ 
   INC$^{\dagger}$ \cite{incrementer} & ViT & 82.4      & 64.2      & 82.2     & 81.6      & 62.2      & 77.6      & 81.4      & 57.1      & 76.2   \\

   CoGaMiD \cite{zhucontinual} & Swin-B & 82.5 & 67.2      & 81.8  & 82.4      & 61.8      & 77.5      & 82.2 & 57.9      & 76.4  \\
   
   MBS$^{\ddagger}$ \cite{DBLP:journals/corr/abs-2407-11859} & ViT &  81.5   & 65.0     & 80.7    &   80.5    & 64.1      &  76.4   &  79.6     & 57.8      & 74.1  \\ \midrule

   
   Ours & ViT  &
   81.8    &  71.1     &  81.9   &  82.2     &  67.9     & 78.6   & 80.6   &   61.5     &   76.6    \\ \midrule\bottomrule  
\end{tabular}
}
\end{table*}

%% file: Tables/ade_complete.tex
\begin{table*}[htb]
\centering
\caption{\small Performance comparison on ADE20K under different scenarios for \textit{overlapped} setting. $\ddagger$ implies results are reproduced from the official repository. $\dagger$ indicates results are excerpted from \cite{DBLP:journals/corr/abs-2407-11859, incrementer}.}
\label{tab:pascal_ade}
\vspace{10pt}
\setlength{\tabcolsep}{3pt}
\renewcommand{\arraystretch}{1}
\resizebox{\linewidth}{!}{
\centering
\begin{tabular}{lc ccc ccc ccc ccc}
\toprule \midrule
\multirow{2}{*}{Methods} & \multirow{2}{*}{Backbone} & \multicolumn{3}{c}{100-50 (2 tasks)} & \multicolumn{3}{c}{50-50 (3 tasks)} 
& \multicolumn{3}{c}{100-10 (6 tasks)} & \multicolumn{3}{c}{100-5 (11 tasks)} \\
\cmidrule(lr){3-5} \cmidrule(lr){6-8} \cmidrule(lr){9-11} \cmidrule(lr){12-14}
& & 1-100 & 101-150 & All & 1-50 & 51-150 & All & 1-100 & 101-150 & All & 1-100 & 101-150 & All \\
\midrule
   MiB$^{\dagger}$ \cite{DBLP:conf/cvpr/CermelliMB0C20} & Res101  & 40.5      & 17.2      & 32.8     & 45.5      & 21.0      & 29.3     & 38.2      & 11.1      & 29.2    & 36.0      & 5.7      & 26.0 \\
   SSUL \cite{DBLP:conf/nips/ChaKYM21} & Res101 & 42.8 & 17.5 & 34.5 & 49.1 & 20.1 & 29.8 & 42.9 & 17.7 & 34.5  & 42.9 & 17.8 & 34.6 \\
   SPPA \cite{DBLP:conf/eccv/LinWZ22} & Res101 & 42.9   & 19.9  & 35.2 & 49.8  & 23.9   & 32.5 & 41.0  & 12.5  & 31.5 & - & - & -  \\
   RCIL$^{\dagger}$ \cite{DBLP:conf/cvpr/ZhangXLCC22}  & Res101  & 42.3      & 18.8      & 34.5     &  48.3     & 25.0      & 32.5     & 39.3      & 17.6      & 32.0   & 38.5      & 11.5      & 29.6    \\
   IDEC \cite{DBLP:journals/pami/ZhaoYS23}& Res101 & 42.0   & 18.2  & 34.1 & 47.4  & 26.0  & 33.1 & 40.3  & 17.6  & 32.7 & 39.2 &  14.6  & 31.0 \\
   LGKD+PLOP \cite{DBLP:conf/iccv/0002LLZWHMHL23} & Res101  & 43.6 &  25.7   & 37.5 & 49.4  & 29.4  & 36.0 & 42.1   & 22.0  & 35.4 & - & - & - \\
   STAR \cite{DBLP:conf/nips/ChenCLIK23} & Res101 & 42.4  & 24.2   & 36.4 & 48.7  & 27.2  & 34.4 & 42.0  & 20.6  & 34.9 & - & - & -  \\
   REMINDER \cite{DBLP:conf/cvpr/PhanTPTB22} & Res101 & 41.6  & 19.2   & 34.1 & 47.1  & 20.4  & 29.4 & 39.0 & 21.3 & 33.1  & 36.1  & 16.4  & 29.5   \\
   PLOP + NeST \cite{DBLP:conf/eccv/XieLXWZL24} & Res101 & 42.2      & 24.3      & 36.3     & 48.7      & 27.7      & 34.8     & 40.9      & 22.0      & 34.7   & 39.3      & 17.4      & 32.0 \\ 
   RCIL + NeST \cite{DBLP:conf/eccv/XieLXWZL24} & Res101  & 42.3      & 22.8      & 35.8     & 48.2      & 27.4      & 34.4     & 40.7      & 19.0      & 33.5   & 39.4      & 15.5      & 31.5  \\
   BARM \cite{DBLP:journals/corr/abs-2407-09838} & Res101  & 42.0      & 23.0      &  35.7    & 47.9      & 26.5      & 33.7     & 41.1      & 23.1      &  35.2  & 40.5      & 21.2      & 34.1 \\
   ADAPTER \cite{DBLP:journals/corr/abs-2412-12669} & Res101  & 43.1  & 23.6  & 36.7 & 49.3   & 27.3  & 34.7 &42.9    & 19.9  & 35.3  & 42.6  & 18.0  & 34.5 \\
   ECLIPSE \cite{Kim_2024_CVPR} & Res101  &  45.0    &  21.7  &  37.1    &  -       &  -     &  -   &  43.4    &  17.4     &  34.6   &  43.3     &  16.3    &  34.2     \\
   CoMBO \cite{fang2025combo} & Res101  &  50.2    &  34.4  &  44.9    &  55.3       &  36.9     &  43.0   &  47.8    &  27.7     &  41.1   &  44.6     &  22.6    &  37.3     \\
   SimCIS \cite{zhu2025rethinking} & Res101  &  57.1    & 36.0 & 48.6   &  -   & -  &  -  &  49.7   & 27.4   &42.3 &  46.7    & 22.8    & 38.7     \\
   EIR \cite{yin2025beyond} & Res101 & 41.9 & 21.9  & 35.3 & 49.3 & 26.3  & 34.1 & 41.8 & 19.4  & 34.3 & - & -  & - \\
   CoGaMiD \cite{zhucontinual} & Res101  &  43.1    &  24.7  &  37.0    &  49.3       &  27.8     &  35.1   &  42.5    &  22.4     &  35.8   &  42.3     &  21.0    &  35.2     \\
   MiB$^{\dagger}$ \cite{DBLP:conf/cvpr/CermelliMB0C20} &  ViT &  46.4     & 35.0      & 42.6     & 52.2      & 35.6      & 41.1    & 43.0      & 30.8      & 38.9     & 40.2      & 26.6      & 35.7   \\
   CoinSeg \cite{Zhang_2023_ICCV} & Swin-B  &41.6   &26.7   & 36.6 & 49.0  & 28.9  & 35.6 & 42.1   & 24.5  & 36.2  & 43.1  & 24.1   &  36.8\\
   INC$^{\dagger}$ \cite{incrementer} & ViT  & 49.4      & 35.6      & 44.8     & 56.2     & 37.8      & 43.9     & 48.5      & 34.6      & 43.9   & 46.9      & 31.3      & 41.7   \\
   MiB + NeST \cite{DBLP:conf/eccv/XieLXWZL24} & Swin-B  &  42.8     & 27.8      & 37.9     & 49.7      & 29.3      & 36.2     & 41.8      & 23.8      & 35.9   & 40.5      & 19.9      & 33.7 \\ 
   PLOP + NeST \cite{DBLP:conf/eccv/XieLXWZL24} & Swin-B  & 43.5      & 26.5      & 37.9     & 50.6      & 28.9      & 36.2     & 41.7      &  24.2     & 35.9   &  39.7     & 18.3      & 32.6   \\
   EIR \cite{yin2025beyond} & Swin-B & 42.1 & 27.3  & 37.2 & 49.7 & 28.8  & 35.8 & 42.3 & 23.6  & 36.1 & - & -  & - \\
   CoGaMiD \cite{zhucontinual} & Swin-B  &  43.9    &  27.3  &  38.4    &  49.9       &  29.8     &  36.6   &  43.7    &  26.5     &  38.0   &  43.6     &  25.8    &  37.7     \\
   
   MBS$^{\ddagger}$ \cite{DBLP:journals/corr/abs-2407-11859} & ViT  &  47.7     & 35.6      & 43.7     & 54.4      & 37.1      &  42.9    & 47.7      & 31.5      &  42.3  &  44.4     &   22.1    & 38.8    \\ \midrule

   Ours & ViT & 54.1  & 40.3   &  48.2   &  56.4   & 42.3   & 47.1   & 50.4  & 35.7   & 44.1   & 46.3      & 29.1      &  41.4  \\ \midrule \bottomrule  
\end{tabular}
}
\end{table*}

%% file: Tables/pascal_cnn.tex
\begin{table*}[ht]
\centering
\caption{\small Performance comparison on Pascal VOC under CNN-based backbone for \textit{overlapped} setting. Best results are marked in \textbf{Bold}.}
\label{tab:cnn}
\vspace{10pt}
\setlength{\tabcolsep}{4pt}
\renewcommand{\arraystretch}{1}
\resizebox{.7\linewidth}{!}{
\centering
\begin{tabular}{lcccccc}
\toprule \midrule
\multirow{2}{*}{Methods} & \multicolumn{3}{c}{15-5 (2 tasks)} & \multicolumn{3}{c}{15-1 (6 tasks)}                        \\ 
\cmidrule(lr){2-4} \cmidrule(lr){5-7}
                 
           &  1-15     & 16-20      & All    &  1-15     & 16-20      & All  \\ \midrule
\multicolumn{7}{c}{\textbf{MiB}}   \\ \midrule
   Baseline \cite{DBLP:conf/cvpr/CermelliMB0C20} & 76.8  & 49.1  & 70.2  &  45.2  &  15.7   & 38.2  \\   
   MiB + NeST \cite{DBLP:conf/eccv/XieLXWZL24}  &  77.1 & 50.1 & 70.7       & 61.7   & \textbf{20.4 } & 51.8  \\ 
   MiB + Ours &  \textbf{ 78.7 }   & \textbf{ 52.1  }   &  \textbf{71.2}     &  \textbf{ 65.8 }   &   18.7    & \textbf{60.6}  \\  \midrule
\multicolumn{7}{c}{\textbf{PLOP}}      \\ \midrule
   Baseline \cite{DBLP:conf/cvpr/DouillardCDC21} &  77.0     &  50.9 &  70.8  &  66.8    & 22.3       &  56.2    \\ 
   PLOP + NeST \cite{DBLP:conf/eccv/XieLXWZL24}  & 77.6     &   55.8       & 72.4  &  72.2   &\textbf{ 33.7 } & 63.1  \\ 
   PLOP + Ours   &\textbf{ 78.1}    & \textbf{  57.2}  &  \textbf{75.3} &  \textbf{ 75.1 }   &  32.4  &   \textbf{70.3}  \\   \midrule
\bottomrule  
\end{tabular}
}
\end{table*}

%% file: Tables/nest_vit.tex
\begin{table*}[ht]
\centering
\caption{\small Performance comparison on Pascal VOC under transformer-based backbones for \textit{overlapped} setting. Best results are marked in \textbf{Bold}.}
\label{tab:nest_vit}
\vspace{10pt}
\setlength{\tabcolsep}{4pt}
\renewcommand{\arraystretch}{1}
\resizebox{.5\linewidth}{!}{
\centering
\begin{tabular}{lcccc}
\toprule \midrule
Models & Backbone & 15-1  & 15-5 & 10-1  \\ \midrule
   Incrementer & ViT-B      & 75.5      &  79.9     & 70.2     \\   
   MiB  & ViT-B    &  53.5     & 80.2      & 25.5    \\ 
   MiB + NeST  & ViT-B     & 76.5      & 80.3      & 71.9     \\  \midrule
   Ours  & ViT-B    & \textbf{80.5}      & \textbf{81.6}      & \textbf{75.7}    \\ \midrule\bottomrule  
\end{tabular}
}
\end{table*}